\documentclass[runningheads]{llncs}

\usepackage{eccv}
\usepackage{eccvabbrv}

\usepackage{graphicx}
\usepackage{booktabs}
\usepackage[accsupp]{axessibility}
\usepackage[table,xcdraw]{xcolor}

\usepackage[breaklinks,colorlinks,urlcolor=eccvblue,citecolor=eccvblue]{hyperref}
\usepackage{tikz}
\usetikzlibrary{positioning,arrows.meta,fit,calc}
\usepackage[utf8]{inputenc}
\usepackage[T1]{fontenc}
\usepackage{url}
\usepackage{booktabs}
\usepackage{amsfonts}
\usepackage{nicefrac}
\usepackage{microtype}
\usepackage{amsmath} 
\usepackage{graphicx}
\usepackage{array}
\usepackage{booktabs}
\usepackage[normalem]{ulem}
\useunder{\uline}{\ul}{}
\usepackage{comment}
\usepackage{wasysym}
\usepackage{multirow}
\usepackage{subcaption}
\usepackage{amssymb}
\usepackage{makecell}
\usepackage{pifont}
\newcommand{\cmk}{\ding{51}}
\newcommand{\xmk}{\ding{55}}

\definecolor{road}{rgb}{0.502,0.000,0.502}
\definecolor{walkway}{rgb}{0.800,0.639,0.282}
\definecolor{dirt}{rgb}{0.502,0.000,0.000}
\definecolor{gravel}{rgb}{0.753,0.753,0.753}
\definecolor{rock}{rgb}{0.965,0.471,0.157}
\definecolor{grass}{rgb}{0.000,1.000,0.000}
\definecolor{vegetation}{rgb}{0.439,0.580,0.125}
\definecolor{tree}{rgb}{0.251,0.251,0.000}
\definecolor{groundobstacles}{rgb}{1.000,1.000,0.000}
\definecolor{person}{rgb}{1.000,0.063,1.000}
\definecolor{bicycle}{rgb}{1.000,0.800,0.600}
\definecolor{vehicle}{rgb}{0.000,0.502,0.502}
\definecolor{water}{rgb}{0.000,0.000,1.000}
\definecolor{building}{rgb}{1.000,0.000,0.000}
\definecolor{roof}{rgb}{0.251,0.627,0.471}
\definecolor{cables}{rgb}{1.000,0.627,0.000}
\definecolor{cabletower}{rgb}{0.416,0.000,1.000}
\definecolor{flyinganimals}{rgb}{0.961,0.592,0.412}
\definecolor{truck}{rgb}{0.500,0.500,0.250}
\definecolor{parkinglot}{rgb}{0.502,0.251,0.502}
\definecolor{constructions}{rgb}{0.941,0.471,0.471}
\definecolor{cranes}{rgb}{1.000,1.000,0.502}

\makeatletter
\newcommand{\road@occuflyfreq}{0.9377}
\newcommand{\walkway@occuflyfreq}{1.0560}
\newcommand{\dirt@occuflyfreq}{1.2102}
\newcommand{\gravel@occuflyfreq}{0.7960}
\newcommand{\rock@occuflyfreq}{0.0210}
\newcommand{\grass@occuflyfreq}{4.1978}
\newcommand{\vegetation@occuflyfreq}{2.3234}
\newcommand{\tree@occuflyfreq}{6.8357}
\newcommand{\groundobstacles@occuflyfreq}{1.7566}
\newcommand{\person@occuflyfreq}{0.0002}
\newcommand{\bicycle@occuflyfreq}{0.0046}
\newcommand{\vehicle@occuflyfreq}{0.5195}
\newcommand{\water@occuflyfreq}{1.1322}
\newcommand{\building@occuflyfreq}{75.7457}
\newcommand{\roof@occuflyfreq}{1.7417}
\newcommand{\cables@occuflyfreq}{0.0017}
\newcommand{\cabletower@occuflyfreq}{0.0040}
\newcommand{\flyinganimals@occuflyfreq}{0.0001}
\newcommand{\parkinglot@occuflyfreq}{1.4349}
\newcommand{\constructions@occuflyfreq}{0.1689}
\newcommand{\cranes@occuflyfreq}{0.0052}
\newcommand{\truck@occuflyfreq}{0.1067}
\newcommand{\occuflyfreq}[1]{{\csname #1@occuflyfreq\endcsname}}
\makeatother

\usepackage{cuted}
\usepackage{multirow}
\usepackage{textcomp}
\usepackage[group-separator={,}, output-decimal-marker={.}, group-minimum-digits=4, group-digits=integer]{siunitx}
\usepackage{orcidlink}

\begin{document}

\title{SegFly: A Dataset and 2D-3D-2D Paradigm\\for Aerial RGB-Thermal Semantic Segmentation\\at Scale} 

\titlerunning{SegFly}

\author{Markus Gross\textsuperscript{1,2,3,$\star$}, Sai B. Matha\textsuperscript{1}, Rui Song\textsuperscript{4,5}, Viswanathan Muthuveerappan\textsuperscript{1}\\ Conrad Christoph\textsuperscript{1}, Julius Huber\textsuperscript{1}, Daniel Cremers\textsuperscript{2,3}
}

\authorrunning{M.~Gross et al.}

\institute{\textsuperscript{1}Fraunhofer Institute IVI \quad \textsuperscript{2}TU Munich \quad \textsuperscript{3}MCML \quad \textsuperscript{4}UCLA \quad \textsuperscript{5}Uni Cambridge}

\maketitle

\begingroup
\renewcommand{\thefootnote}{}
\footnotetext{$^\star$Corresponding author: markus.gross@tum.de.}
\endgroup

\begin{abstract}
    Semantic segmentation for uncrewed aerial vehicles (UAVs) is fundamental for aerial scene understanding, yet existing RGB and \mbox{RGB-T} datasets remain limited in scale, diversity, and annotation efficiency due to the high cost of manual labeling and the difficulties of accurate RGB-T alignment on off-the-shelf UAVs. To address these challenges, we propose a scalable geometry-driven 2D-3D-2D paradigm that leverages multi-view redundancy in high-overlap aerial imagery to automatically propagate labels from a small subset of manually annotated RGB images to both RGB and thermal modalities within a unified framework. By lifting less than 3\% of RGB images into a semantic 3D point cloud and rendering it into all views, our approach enables dense pseudo ground-truth generation across large image collections, automatically producing 97\% of RGB labels and 100\% of thermal labels while achieving 91\% and 88\% annotation accuracy without any 2D manual refinement.
    We further extend this 2D–3D–2D paradigm to cross-modal image registration, using 3D geometry as an intermediate alignment space to obtain fully automatic, strong pixel-level RGB-T alignment with 87\% registration accuracy and no hardware-level synchronization. Applying our framework to existing geo-referenced aerial imagery, we construct SegFly, a large-scale benchmark with over 20,000 high-resolution RGB images and more than 15,000 geometrically aligned RGB-T pairs spanning diverse urban, industrial, and rural environments across multiple altitudes and seasons. On SegFly, we establish the Firefly baseline for RGB and thermal semantic segmentation and show that both conventional architectures and vision foundation models benefit substantially from SegFly supervision, highlighting the potential of geometry-driven 2D-3D-2D pipelines for scalable multi-modal aerial scene understanding. The SegFly dataset and our Firefly baseline are available at \url{https://github.com/markus-42/SegFly}.
\end{abstract}

\section{Introduction}
\label{sec_intro}

Uncrewed aerial vehicles (UAVs) equipped with visual sensors have become a central platform for large-scale environment perception in computer vision.
Recent advances in aerial vision enable applications such as delivery and logistics, infrastructure inspection, environmental monitoring, emergency response, surveillance, precision agriculture, and urban planning~\cite{farajijalal2025safety, gross2026safeland, sssurvey}.
These systems rely on robust scene understanding from aerial imagery captured under diverse environmental conditions.
While most UAV platforms rely on RGB cameras, thermal sensors provide complementary information that is largely invariant to illumination changes and adverse weather, making RGB-Thermal (RGB-T) perception attractive for real-world deployments~\cite{mahes2026anythermal}.

At the core of aerial perception are fundamental vision tasks including object detection~\cite{dhaouadi2025deepscenarionO3d,meier2025carladrone}, tracking~\cite{sturm2013tracking}, localization~\cite{dhaouadi2025ortholoc}, mapping and navigation~\cite{stumberg2016autonomous}, and 3D reconstruction~\cite{gross2025occufly}.
Among these, semantic segmentation provides dense pixel-level scene understanding that directly supports higher-level reasoning and decision making~\cite{csurka2022semantic_segmentation_history}.
However, large-scale aerial semantic segmentation remains challenging.
Aerial imagery exhibits strong viewpoint variability, large scale changes, and complex scene layouts.
For RGB-T perception, datasets additionally require accurate cross-modal alignment, which is difficult due to asynchronous capture, platform motion, and sensor characteristics.
Existing solutions therefore rely on synchronized hardware or extensive manual refinement.

\begin{figure}[t]
	\centering
	\includegraphics[width=0.99\linewidth]{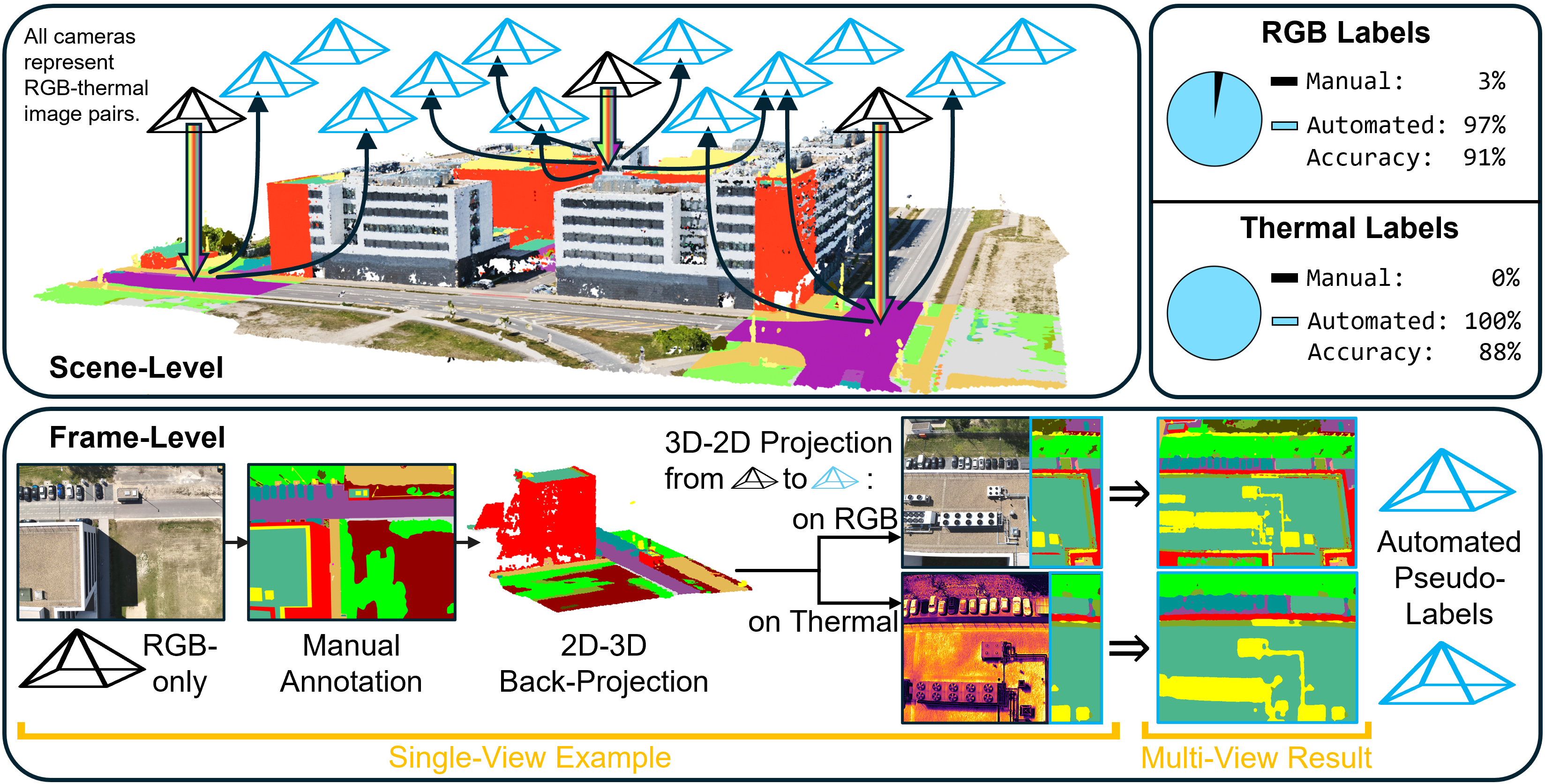}
	\caption{2D-3D-2D Paradigm. An overview is provided in~\cref{subsec_overview}.}
	\label{fig_banner}    
\end{figure}

Another major limitation is the cost of dense pixel-level annotation.
High-resolution aerial imagery contains complex structures that require careful labeling, making large-scale annotation expensive.
As a result, most aerial segmentation datasets remain limited in scale and diversity, particularly for RGB-T data where aligned image pairs are required.
Current datasets typically contain only a few thousand samples, restricting the ability of modern deep learning models to fully benefit from large-scale training data.
Recent work has explored geometry-driven supervision to address these challenges.
The 2D-to-3D label lifting paradigm leverages multi-view imagery to propagate semantic annotations through reconstructed geometry.
The OccuFly benchmark~\cite{gross2025occufly} demonstrated that labels from a small subset of annotated images can be lifted into a 3D point cloud and reused across multiple views.
However, existing approaches focus on RGB imagery and do not address large-scale multi-modal segmentation or RGB-T alignment.

In this work, we present a scalable framework for aerial RGB and thermal semantic segmentation based on a geometry-driven 2D–3D–2D paradigm, shown in \cref{fig_banner}. By annotating only a small subset of RGB images, semantic labels are lifted into a reconstructed 3D point cloud and reprojected into both RGB and thermal views to generate dense pseudo ground truth at scale. Thermal imagery is additionally reconstructed in 3D and aligned with the RGB semantic point cloud, enabling accurate RGB-T registration through geometric alignment and lens distortion transfer entirely in software, eliminating the need for hardware synchronization and enabling scalable data collection with off-the-shelf UAV platforms. Building on this pipeline, we introduce SegFly, a large-scale benchmark for aerial semantic segmentation across RGB and RGB-T modalities, comprising more than 20,000 high-resolution RGB images and over 15,000 geometrically aligned RGB-T pairs captured across diverse environments, seasons, and flight altitudes. This scale is enabled by geometry-driven pseudo-label generation, requiring manual annotation for less than 3\% of the collected data. Using SegFly, we establish baseline results with the Firefly architecture, providing a strong reference for both RGB and RGB-T segmentation. Furthermore, we show that RGB and RGB-T vision foundation models benefit substantially from training on SegFly, highlighting the effectiveness of large-scale geometry-derived supervision for multi-modal aerial perception.

\vspace{0.5em}
\noindent In summary, our contributions are as follows:

\begin{itemize}
    \item We propose a scalable \textbf{2D-3D-2D framework} that lifts manual semantic annotations from  <\SI{3}{\%} of RGB images into a reconstructed 3D point cloud and renders dense pseudo ground-truth across \SI{97}{\%} of RGB and \SI{100}{\%} of thermal views, achieving \SI{91}{\%} and \SI{88}{\%} accuracy without human refinement.
    \item Based on our framework, we present \textbf{SegFly, a large-scale benchmark} for aerial semantic segmentation providing >$20,000$ RGB samples and >$15,000$ aligned RGB-T samples across diverse environments, seasons, and altitudes.
    \item Following the 2D–3D–2D paradigm, we propose a \textbf{multi-modal geometry-driven RGB-T registration method}, achieving \SI{87.05}{\%} pixel-level accuracy, thereby eliminating hardware synchronization. This enables practical and scalable data collection with off-the-shelf multi-modal UAVs, significantly reducing barriers to the construction of large-scale RGB-T datasets.
    \item On SegFly, we establish the \textbf{Firefly baseline}, a simple yet powerful and efficient method for aerial RGB and thermal semantic segmentation.
    \item We further show that RGB and RGB-T \textbf{aerial vision foundation models} benefit substantially from SegFly supervision, highlighting the potential to further reduce manual annotation toward fully automated data generation.
\end{itemize}

\begin{table}[t!] 
 \centering 
 \caption{SegFly in comparison to established aerial RGB and RGB-T datasets.} 
 \label{tab_semantic_datasets} 
 \small 
 \renewcommand{\arraystretch}{1.2} 
 \resizebox{\linewidth}{!}{ 
 \begin{tabular}{l|c|c|c|c|c|r|c|c|ccc|l} 
 \toprule 
 \multirow{2}{*}{Dataset} & 
 \multirow{2}{*}{Modality} & 
 RGB-T & 
 Resolution & 
 Resolution & 
 Altitude & 
 \multicolumn{1}{c|}{\# of} & 
 \# of & 
 Annotation & 
 \multicolumn{3}{c|}{Environment} & 
 \multicolumn{1}{c}{\multirow{2}{*}{Venue}} \\
 & & Aligned & RGB & Thermal & [m] &
 \multicolumn{1}{c|}{Samples} & Classes & Density &
 Urban & Industrial & Rural & \\
 \midrule 
 \multicolumn{13}{l}{\textbf{RGB Datasets}} \\
 \hline
 \rowcolor[HTML]{E6E6E6} 
Aeroscapes~\cite{nigam2018aeroscapes}
   & RGB & - 
   & $1280 $$\times$$ 720$ & - 
   & $5\text{--}30$ 
   & $3,269$ & $12$ & Low & \cmk & & \cmk & WACV '18 \\ 
 
 Okutswiss~\cite{speth2022otukamas}
   & RGB & - 
   & $4608 $$\times$$ 3456$ / $3840 $$\times$$ 2160$ & - 
   & $70; 80$ 
   & $191$ & $10$ & Medium & \cmk & & \cmk & JFR '22 \\ 
 
 \rowcolor[HTML]{E6E6E6} 
 ICG~\cite{icgDataset} 
   & RGB & - 
   & $6000 $$\times$$ 4000$ & - 
   & $5\text{--}50$ 
   & $600$ & $20$ & High & \cmk & & & - \\ 
 
 UAVid~\cite{lyu2020uavid}
   & RGB & - 
   & $4096 $$\times$$ 2160$ / $3840 $$\times$$ 2160$ & - 
   & $50$ 
   & $300$ & $8$ & Medium & \cmk & & & ISPRS JPRS '20 \\ 
 
 \rowcolor[HTML]{E6E6E6} 
 UDD6~\cite{chen2018udd} 
   & RGB & - 
   & $4096 $$\times$$ 2160$ / $4000 $$\times$$ 3000$ & - 
   & $60\text{--}100$ 
   & $141$ & $6$ & Medium & \cmk & & & PRCV '18 \\ 
 
 VDD~\cite{cai2025vdd}
   & RGB & - 
   & $4000 $$\times$$ 3000$ & - 
   & $50\text{--}120$ 
   & $400$ & $7$ & Medium & \cmk & \cmk & \cmk & JVCIR '25 \\ 
 
 \rowcolor[HTML]{E6E6E6} 
 Floodnet~\cite{rahn2021floodnet}
   & RGB & - 
   & $4000 $$\times$$ 3000$ & - 
   & ``low altitude'' 
   & $2,343$ & $10$ & Medium & \cmk & & & IEEE Access '21 \\ 
 
 UAVScenes~\cite{wang2025uavscenes}
   & RGB & - 
   & $2448 $$\times$$ 2048$ & - 
   & $80; 90; 130$ 
   & $120,000$ & $19$ & Medium & & & \cmk & ICCV '25 \\ 
 
 \rowcolor[HTML]{E6E6E6} 
 MESSI~\cite{pinkovich2025messi}
   & RGB & - 
   & $5472 $$\times$$ 3684$ & - 
   & $30\text{--}100$ 
   & $2,525$ & $16$ & High & \cmk & & & TMLR '25 \\ 
 
 CrossLoc~\cite{yan2021crossloc}
   & RGB & - 
   & $720 $$\times$$ 480$ & - 
   & - 
   & $7,000$ & $7$ & High & \cmk & & \cmk & CVPR '22 \\ 
 
 \rowcolor[HTML]{E6E6E6} 
 SkyScapes~\cite{azimi2019skyscapes}
   & RGB & - 
   & $5616 $$\times$$ 3744$ & - 
   & $1,000$ 
   & $16$ & $31$ & High & \cmk & & \cmk & ICCV '19 \\ 
 
 \hline
 SegFly-RGB (ours) 
   & RGB & - 
   & $5472$$\times$$3648$ / $4000 $$\times$$ 3000$ & - 
   & $30; 40; 50$ 
   & $20,606$ & $15$ & High & \cmk & \cmk & \cmk & \\ 
 
 \midrule
 \multicolumn{13}{l}{\textbf{RGB-Thermal Datasets}} \\
 \hline
 Kust4K~\cite{ouyang2025kust4k}
   & RGB+T & \cmk 
   & $640 $$\times$$ 512$ & $640 $$\times$$ 512$ 
   & - 
   & $4,024$ & $8$ & Medium & \cmk & & & Nature Sci-Data '25 \\ 
 
 \rowcolor[HTML]{E6E6E6} CART~\cite{lee2024cart}
   & RGB+T & \cmk 
   & $960 $$\times$$ 600$ & $640 $$\times$$ 512$ 
   & $40$ 
   & $2,282$ & $11$ & High & & & \cmk & ECCV '24 \\ 
 
 MVUAV~\cite{cheng2025mvuav}
   & RGB+T & \cmk 
   & $1920 $$\times$$ 1080$ & - 
   & $5\text{--}20$ 
   & $2,183$ & $36$ & High & \cmk & & & NeurIPS '24 \\ 
 
 \rowcolor[HTML]{E6E6E6} 
 IndraEye~\cite{manjunath2025indraeye}
   & RGB+T & \xmk 
   & $1280 $$\times$$ 720$ & $640 $$\times$$ 480$ 
   & $7\text{--}30$ 
   & $5,612$ & $13$ & Low & \cmk & & & CVPRW '25 \\ 
 
 \hline
 SegFly-RGB-T (ours) 
   & RGB+T & \cmk 
   & $640 $$\times$$ 512$ & $640 $$\times$$ 512$ 
   & $30; 40; 50$ 
   & $15,007$ & $15$ & High & \cmk & \cmk & \cmk &  \\ 
 
 \bottomrule 
 \end{tabular} 
 } 
\end{table}

\section{Related Work}
\label{sec_related_work}

\noindent\textbf{Datasets.}\;
Semantic segmentation from aerial platforms has a long history, which is reflected in established RGB datasets~\cite{pinkovich2025messi, azimi2019skyscapes, yan2021crossloc, wang2025uavscenes, cai2025vdd, chen2018udd, lyu2020uavid, speth2022otukamas, nigam2018aeroscapes, rahn2021floodnet} and RGB-T datasets~\cite{ouyang2025kust4k, lee2024cart,cheng2025mvuav, manjunath2025indraeye}, summarized in ~\cref{tab_semantic_datasets}.
Early works~\cite{nigam2018aeroscapes, lyu2020uavid, chen2018udd} provide high-resolution imagery with pixel-wise labels, but are limited to a few hundred images and a small number of classes, while more recent datasets \cite{wang2025uavscenes, yan2021crossloc, ouyang2025kust4k, lee2024cart} push toward higher class diversity and overall scale.
Notably, RGB and RGB-T domains share the same bottleneck:
they rely on manual 2D annotation as the primary source of supervision, resulting in costly and time-consuming human effort that limits scalability.
To mitigate this bottleneck, existing work renders synthetic data~\cite{yan2021crossloc}, projects 3D labels with non‑trivial manual refinement \cite{wang2025uavscenes}, or exploits foundation models to partially automate 2D masks~\cite{ouyang2025kust4k}.
However, existing approaches remain inefficient, as they still require substantial manual effort, including hundreds of man-hours, to maintain consistency across large numbers of small and thin structures~\cite{wang2025uavscenes,nigam2018aeroscapes}.
Furthermore, for the RGB-T domain, existing methods inherit strong assumptions about data acquisition to ensure aligned RGB and thermal images that are necessary for improved segmentation results.
Specifically, they rely on custom, hardware‑synchronized sensor rigs with static extrinsics~\cite{lee2024cart,cheng2025mvuav,mahes2026anythermal}, or perform SIFT‑based registration and manual cropping~\cite{ouyang2025kust4k}.
As a result, registration quality and scalability are tied to specific hardware configurations or substantial manual effort. 

In contrast, SegFly is designed from the outset for off‑the‑shelf multi‑modal UAVs that lack hardware‑level time synchronization. 
By reconstructing RGB and thermal imagery in 3D, aligning their point clouds, and then transferring lens distortion back to 2D, our 2D–3D–2D pipeline achieves geometry‑guided, precise RGB‑T alignment entirely in software.
Beyond RGB-T registration, we consider recent insights in 3D aerial semantic vision~\cite{gross2025occufly} to utilize our 2D-3D-2D approach for semantic labeling of both modalities: instead of annotating every frame independently, we exploit the multi-view redundancy and metric accuracy of high-overlap, geo-referenced RGB-T imagery, by annotating a small subset of \emph{RGB-only} images in 2D (<\SI{3}{\%}), lift them into a semantic 3D point cloud, and render it back into the remaining >\SI{97}{\%} of RGB views and \SI{100}{\%} of thermal views.\\

\noindent\textbf{Methods.}\;
Apart from established deep semantic segmentation architectures such as UPerNet~\cite{xiao2018upernet} and SegFormer~\cite{xiao2018upernet}, recent progress in the RGB domain has shifted toward leveraging vision foundation models~\cite{simeoni2025dinov3} for generalizable dense prediction, often combined with parameter‑efficient fine‑tuning~\cite{chen2022adapter,hu2022lora,wei2024rein}.
Vision-based segmentation foundation models, such as the SegmentAnythingModel~\cite{kirillov2023segment} and adaptations such as SemanticSegmentAnything~\cite{chen2023semanticsegmentanything}, provide broad generalization and interactive segmentation capabilities beyond fixed taxonomies.
However, despite these advances, there remains tangible scope for improvement in aerial domains~\cite{prado2023sam}, where unique challenges such as extreme scale variation and viewpoint geometry degrade performance.
To address these challenges in the aerial RGB domain, approaches like CatSeg~\cite{cho2024catseg} and SegEarth‑OV~\cite{li2025segearthov} introduce open‑vocabulary or cost‑aggregation mechanisms to improve segmentation generality, while in the RGB‑T domain, models such as AnyThermal~\cite{mahes2026anythermal} learn universal thermal representations via distillation from visual foundation models to better support thermal segmentation.
We evaluate these models in a zero-shot setting and find that their performance on aerial imagery with the complexity present in SegFly is limited.
However, when fine‑tuned with SegFly, their performance improves substantially, highlighting the potential to further reduce manual annotation toward fully automated 2D and 3D aerial semantic data generation.

\section{Data Generation Framework}
\label{sec_data_generation}

\subsection{Overview}
\label{subsec_overview}

Our data generation pipeline proposes a scalable \emph{2D-3D-2D} paradigm that exploits the inherent multi-view redundancy of high-overlap aerial imagery to produce dense, high-quality semantic pseudo-labels for both RGB and thermal images at unprecedented scale.
First, we perform classical multi-view geometry on the RGB images alone, yielding a metric 3D point cloud together with dense 2D--3D correspondences.
In line with OccuFly~\cite{gross2025occufly}, labels are then lifted from a small subset of manually annotated RGB images into this 3D point cloud, producing a semantically labeled point cloud with minimal human effort (see~\cref{subsubsec_2d_3d}).

In the novel 3D--2D stage, we project the semantic point cloud onto \emph{every} target frame, both the remaining RGB images and their corresponding thermal images. 
To address sparsity and occlusion artifacts inherent to point-based projection, we apply a four-step depth-guided semantic projection process: (1) semantic projection with Z-buffering, (2) depth-aware occlusion filtering, (3) splatting-based densification, and (4) depth-guided label propagation.
The result is a dense, high-fidelity pseudo-label map for both RGB and thermal modalities, for which we annotate only \SI{3}{\%} of the RGB images manually while automatically generating \SI{97}{\%} of RGB and \SI{100}{\%} of thermal pseudo-labels.

Finally, to enable cross-modal domain adaptation, RGB-T image pairs must be registered. Rather than relying on cross-modal 2D feature matching or learnable homographies, we extend the 2D–3D–2D paradigm to cross-modal geometry-guided registration: thermal images are first reconstructed in 3D, aligned to the RGB point cloud using ICP, and then projected back to 2D via a lens-distortion transfer.
This process achieves high-fidelity, strong pixel-level registration without hardware-level time-synchronization or strictly fixed extrinsics, enabling scalable data acquisition on off-the-shelf UAVs (see~\cref{subsec_rgbt_registration}).

\subsection{Problem Formulation}
\label{subsec_problem_formulation}

Let $\mathcal{I}^\text{RGB} = \{\mathbf{I}_n^\text{RGB} \in \mathbb{R}^{H \times W \times 3}\}_{n=1}^N$ and $\mathcal{I}^\text{Th} = \{\mathbf{I}_n^\text{Th} \in \mathbb{R}^{H \times W \times 1}\}_{n=1}^N$ denote the sets of RGB and thermal \emph{image pairs}, respectively, acquired by a multi-modal UAV.
All images have per-modality, geo-referenced world-to-camera poses \mbox{$\mathcal{T}^\text{RGB/Th} = \{\mathbf{T}_n^{c\leftarrow w} \in \text{SE}(3)\}_{n=1}^N$} and intrinsics $\mathcal{K}^\text{RGB/Th} \in \mathbb{R}^{3\times3}$.
We define a semantic label set $\mathcal{C} = \{1, \dots, C\}$, and manual pixel-wise annotations are provided only for a small subset of RGB source images indexed by 
$\mathcal{S}\subset\{1,\dots,S\}$ with $|\mathcal{S}| \ll N$.
The remaining images $\mathcal{U}=\{1,\dots,N\}\setminus\mathcal{S}$ are treated as target views.
Our objective is to generate dense semantic annotations
\begin{equation}
    \mathbf{Y}_n^{\text{RGB}},\;
    \mathbf{Y}_n^{\text{Th}}
    :\{1,\dots,H\}\times\{1,\dots,W\}\rightarrow\mathcal{C},
    \quad \forall n\in\mathcal{U},
\end{equation}
for both RGB and thermal images, by exploiting a semantically labeled 3D point cloud $\mathcal{P}_{\mathcal{S}}$ reconstructed from the full RGB image set $\mathcal{I}^\text{RGB}$ and annotated only from the small subset of RGB source images $\mathcal{S}$.

Furthermore, in the context of cross-modal domain adaptation, for every thermal image $\mathcal{I}^{\text{Th}}_n$ we aim to register its corresponding RGB image $\mathcal{I}^{\text{RGB}}_n$, resulting in registered pairs $(\mathcal{I}^{\text{Th}}_n, \mathcal{I}^{\text{RGB,reg}}_n)$.

\begin{figure}[t!]
	\centering
	\includegraphics[width=1.0\linewidth]{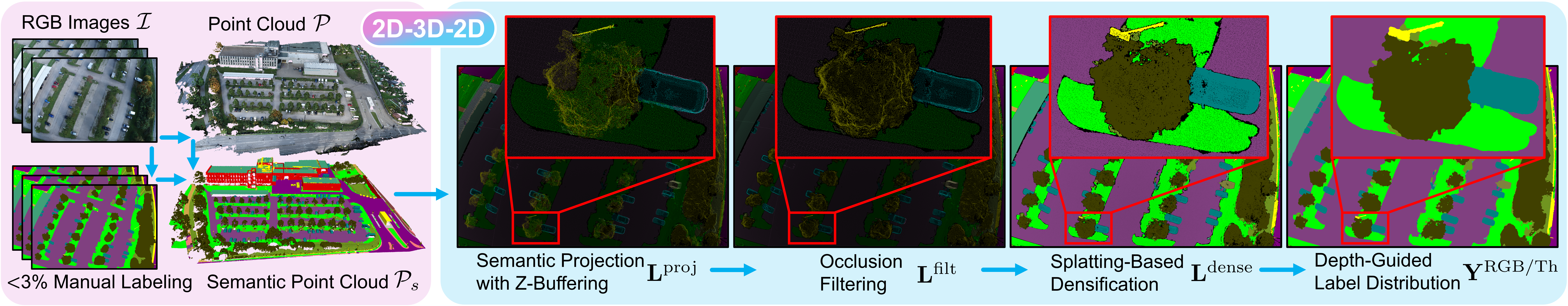}
	\caption{Proposed 3D-to-2D point-based semantic rendering pipeline (\cref{subsubsec_3d_2d}).}
	\label{fig_2d3d2d}    
\end{figure}

\subsection{2D--3D--2D Pseudo Ground-Truth Generation}
\label{subsec_2d3d2d}

\subsubsection{2D-to-3D Stage:}
\label{subsubsec_2d_3d}

This stage directly follows the semantic 3D reconstruction pipeline of OccuFly~\cite{gross2025occufly}, illustrated in \cref{fig_2d3d2d}.
To summarize, using Structure-from-Motion (SfM) and Multi-View Stereo (MVS), we reconstruct a dense metric point cloud from all RGB images:
\begin{equation}
(\mathcal{P}, \boldsymbol{\kappa}^\text{RGB}, \mathcal{A}) 
= \Psi_{\text{SfM+MVS}}(\mathcal{I}^\text{RGB}, \mathcal{K}^\text{RGB}, \mathcal{T}^\text{RGB}),
\end{equation}
where $\mathcal{P} = \{\mathbf{x}_m \in \mathbb{R}^3\}_{m=1}^M$ is the set of reconstructed 3D points, and $\boldsymbol{\kappa}^\text{RGB}$ the estimated distortion coefficients. $\mathcal{A}_n$ are 2D–3D correspondences between a 3D point $\mathbf{x}$ and its 2D projection to pixel location $(u,v)$ satisfying
\begin{equation}
((u,v),\mathbf{x})\in\mathcal{A}_n
\iff
(u,v,1)^\top\sim
\mathbf{K}_n[\,\mathbf{R}_n\,|\,\mathbf{t}_n\,]
[\mathbf{x}^\top\ 1]^\top.
\end{equation}

To minimize manual effort, we annotate only a small spatially stratified subset $\mathcal{S} \subset \{1,\dots,S\}$ of the RGB images, selected by partitioning the ground plane into \SI{25}{m} cells and choosing the closest center-pose per cell.
This selection empirically ensures that over 99\% of the reconstructed points are observed by at least one annotated image, while \(|\mathcal{S}|/N < 0.1\) on average, quantified by the coverage metric
\begin{equation}
\frac{\bigl|\{\mathbf{x} \in \mathcal{P} \mid \exists n \in \mathcal{S},\, \exists (u,v)\ \text{s.t.}\ ((u,v),\mathbf{x}) \in \mathcal{A}_n \}\bigr|}{|\mathcal{P}|} > 0.99.
\end{equation}
After manual per-pixel annotation of images in $\mathcal{S}$, labels are lifted to 3D via back-projection using the correspondences $\mathcal{A}$.
As most of the points are observed by multiple cameras, we fuse multi-view evidence by unweighted majority voting, where ties are broken by a fixed class-frequency prior.
Unlabeled points are completed with inverse-distance-weighted kNN, followed by a second denoising kNN pass, analogous to~\cite{tian2023occ3d}.
The output is a semantically annotated point cloud

\begin{equation}
\mathcal{P}_{\mathcal{S}}=
\{(\mathbf{x}_m,c_m)\}_{m=1}^{M},
\quad c_m\in\mathcal{C}.
\end{equation}

\subsubsection{3D-to-2D Stage:}
\label{subsubsec_3d_2d}

We now render the semantic point cloud $\mathcal{P}_{\mathcal{S}}$ into all target views $n\in\mathcal{U}$ to obtain dense pseudo-labels $\mathbf{Y}_n^{\text{RGB/Th}}$ for both RGB and thermal images, illustrated in \cref{fig_2d3d2d}.
Our rendering pipeline consists of four stages:

\paragraph{(1) Semantic Projection with Z-Buffering.} For each target frame $n \in \mathcal{U}$ (RGB and thermal), we project every point $\mathbf{x}_m \in \mathcal{P}_\mathcal{S}$ into the image plane using the respective camera model.
The projection includes full radial and tangential distortion correction, with RGB distortion coefficients $\boldsymbol{\kappa}^\text{RGB}$ and thermal coefficients $\boldsymbol{\kappa}^\text{Th}$ (discussed in \cref{subsec_rgbt_registration}).
Moreover, we maintain a per-pixel Z-buffer to resolve depth conflicts, retaining only the point with smallest depth $D$ at each pixel $(u,v)$.
This yields an initial sparse label map $\mathbf{L}_n^\text{proj} \in (\mathcal{C} \cup \{0\})^{H \times W}$, where $0$ denotes no projection.

\paragraph{(2) Occlusion Filtering.}
Projected points in the sparse label map $\mathbf{L}_n^\text{proj}$ that survive Z-buffering may still be geometrically occluded by closer geometry of a different semantic class.
We remove such inconsistent projections via a local kernel-based visibility filter.
For each labeled pixel $(u,v)$ with label $c$ and depth $D(u,v)$, we consider a local neighborhood $\Omega_{u,v}$ of size $k\times k$.
If there exists $(u',v')\in\Omega_{u,v}$ such that
\[
\mathbf{L}_n^\text{proj}(u',v')\neq c
\quad\text{and}\quad
D_n(u',v') < D_n(u,v)-\tau,
\]
with depth threshold $\tau>0$, 
we set $\mathbf{L}_n^\text{proj}(u,v)=0$ as likely occluded.
We empirically determined $\tau =$~\SI{0.2}{m}, while $\Omega^{\text{RGB}}=9\times9$ and $\Omega^{\text{Th}}=5\times5$, due to lower resolution of thermal images.
This step transforms the initial sparse label $\mathbf{L}_n^\text{proj}$ to a filtered label map $\mathbf{L}_n^\text{filt}$, by removing geometrically inconsistent label projections.

\paragraph{(3) Splatting-based Densification.}
The filtered label map $\mathbf{L}_n^\text{filt}$ remains sparse due to point cloud sampling density.
We increase density by splatting each labeled pixel to a small disk of fixed radius $r_\text{splat}$, increasing spatial coverage while preserving local consistency.
To incorporate occlusion awareness, we process the splats in back-to-front order (decreasing depth), ensuring that nearer labels take precedence in overlapping regions, resulting in the densified label map $\mathbf{L}_n^\text{dense}$.
In practice, $r_\text{splat}^{\text{RGB}}=3$ and $r_\text{splat}^{\text{Th}}=1$.
The core of this step is standard point-splatting~\cite{zwicker2001splat}, extended to incorporate semantics labels and occlusion awareness.

\paragraph{(4) Depth-Guided Label Distribution.}
Remaining unlabeled pixels in $\mathbf{L}_n^\text{dense}$ are filled using a two-pass k-nearest-neighbor (kNN) scheme that explicitly respects depth ordering.
In the first pass, for each unlabeled pixel, we retrieve its $k$ nearest labeled neighbors in 2D image space.
Among these, we assign the label of the neighbor with the smallest depth to the camera center, ensuring foreground priority.
In a final pass, all pixels are refined using standard (depth-agnostic) 2D kNN with a larger neighborhood, ensuring consistent label boundaries.
We empirically select \(k=5\) for the first pass, while for the second pass, $k^{\text{RGB}}$ ranges from 7 to 25 and $k^{\text{Th}}$ from 13 to 15.
The resulting dense label maps $(\mathbf{Y}_n^{\text{RGB}}$, $\mathbf{Y}_n^{\text{Th}})$ constitute the pseudo ground-truth for their corresponding images $(\mathbf{I}_n^\text{RGB},\mathbf{I}_n^\text{Th})$.\newline

\subsection{RGB-Thermal Image Registration}
\label{subsec_rgbt_registration}
For every thermal image \(\mathcal{I}^{\text{Th}}_n\) we aim to register its corresponding RGB image \(\mathcal{I}^{\text{RGB}}_n\), resulting in registered pairs \((\mathcal{I}^{\text{Th}}_n, \mathcal{I}^{\text{RGB,reg}}_n)\).
To accomplish this, necessary RGB intrinsics, distortion coefficients, and poses \((\mathcal{K}^\text{RGB}, \boldsymbol{\kappa}^\text{RGB}, \mathcal{T}^\text{RGB})\) are obtained as described in \cref{subsubsec_2d_3d}. 
Thermal intrinsics and distortion coefficients \((\mathcal{K}^\text{Th}, \boldsymbol{\kappa}^\text{Th})\) are determined via a custom passive calibration target, following~\cite{shivakumar2020pst900}.
Similar to the RGB-based reconstruction in \ref{subsubsec_2d_3d}, we first perform an independent SfM and MVS reconstruction on the thermal images:
\begin{equation}
(\mathcal{P}^\text{Th}, \mathcal{T}^\text{Th,ref})
= \Psi_{\text{SfM+MVS}}(\mathcal{I}^\text{Th}, \mathcal{K}^\text{Th}, \mathcal{T}^\text{Th}),
\end{equation}
yielding a thermal point cloud $\mathcal{P}^\text{Th}$ and globally refined thermal poses $\mathcal{T}^\text{Th,ref}$. 
RGB-T synchronization is then obtained by aligning \(\mathcal{P}^\text{Th}\) to the RGB-based semantic point cloud \(\mathcal{P}_\mathcal{S}\) using the Iterative Closest Point (ICP) algorithm~\cite{besl1992icp}, producing the rigid transformation \(\mathbf{T}^\text{reg} \in \text{SE}(3)\).
The registered thermal poses are given by
\begin{equation}
\mathcal{T}^\text{Th,reg} = \mathbf{T}^\text{reg} \circ \mathcal{T}^\text{Th,rel},
\end{equation}
where \(\circ\) denotes the composition of rigid transformations in SE(3).
This synchronization matches geometric structures across all images without requiring time-synchronization or static sensor rigs.

With registered thermal poses $\mathcal{T}^{\text{Th,reg}}$ available, we formulate registration in a common undistorted pinhole domain shared by both cameras.
Let $\bar{\pi}^{\text{RGB}}$ and $\bar{\pi}^{\text{Th}}$ denote the undistorted projection functions obtained by inverting the RGB and thermal distortion models $\boldsymbol{\kappa}^{\text{RGB}}$ and $\boldsymbol{\kappa}^{\text{Th}}$, respectively.
For each thermal image $n$, we first determine the RGB support region corresponding to the thermal field of view.

To accomplish this, we first project all 3D points $\mathbf{x}_m \in \mathcal{P}_{\mathcal{S}}$ that are visible in the registered thermal camera into the undistorted thermal image using $\bar{\pi}^{\text{Th}}$ and pose $\mathbf{T}_n^{\text{Th,reg}}$, and then reproject the same points into the undistorted RGB image using $\bar{\pi}^{\text{RGB}}$ and $\mathbf{T}_n^{\text{RGB}}$.
The extremal projected coordinates define an axis-aligned bounding box $\bar{\mathcal{B}}_n$ in this common pinhole domain.
We then synthesize a registered RGB image in the thermal distortion space via a single lens-distortion transfer.
Concretely, we resample the original (distorted) RGB image into the undistorted domain using the RGB camera warp (RGB intrinsics $+$ $\boldsymbol{\kappa}^{\text{RGB}}$) and subsequently map this undistorted RGB image into the thermal image plane using the thermal camera warp (thermal intrinsics $+$ $\boldsymbol{\kappa}^{\text{Th}}$).
The same thermal camera warp is applied to the bounding box $\bar{\mathcal{B}}_n$, after which we crop and resize to the thermal resolution.
This yields the registered RGB image $\mathbf{I}_n^{\text{RGB,reg}}$, which is aligned with $\mathbf{I}_n^{\text{Th}}$ in the thermal distortion space, closing the 2D–3D–2D loop for cross-modal registration.

By following the 2D-3D-2D paradigm, the resulting pairs $(\mathbf{I}_n^{\text{Th}}, \mathbf{I}_n^{\text{RGB,reg}})$ achieve highly accurate, strong pixel-level cross-modal consistency without cross-modal deterministic or learnable 2D features, homography estimation, manual correction, or hardware-level time synchronization, enabling high-fidelity registration on off-the-shelf UAVs, thereby streamlining data acquisition for scalable aerial RGB-T semantic segmentation.

\section{Firefly Baseline}
\label{sec_firefly_method}

We introduce Firefly, a simple yet strong and efficient aerial baseline that addresses (1) RGB semantic segmentation, and (2) thermal semantic segmentation within a unified framework (see ~\cref{fig_firefly}).
Our objective is to achieve strong RGB performance while enabling accurate thermal segmentation through RGB domain adaptation that transfers knowledge from the RGB modality to the thermal domain in a parameter-efficient manner. 

Training proceeds in three stages, termed RGB pre-training, RGB-Thermal domain adaptation, and thermal fine-tuning: 
In \emph{\textbf{RGB pre-training}}, a fixed \mbox{DINOv3~\cite{simeoni2025dinov3}} encoder is combined with a lightweight multi-layer perceptron (MLP) head~\cite{rumelhart1986mlp} that is optimized on labeled RGB data to establish a strong baseline for RGB-based semantic segmentation. 
Subsequently, for \emph{\textbf{RGB-Thermal domain adaptation}}, registered RGB and thermal image pairs are leveraged to align feature representations in a self-supervised and parameter-efficient manner.
This is accomplished by utilizing thermal-specific Rein~\cite{wei2024rein} adapters, which learn residual mappings that transform thermal inputs into the RGB feature space, while keeping all other parameters fixed. 
Finally, for \emph{\textbf{thermal fine-tuning}}, a dedicated thermal MLP head initialized from the RGB head is trained jointly with the Rein adapters on labeled thermal data, while the encoder and RGB head remain frozen, resulting in accurate thermal segmentation without compromising RGB performance.
Full technical details of the adapter design, feature-level objectives, and optimization procedure are provided in the supplementary material.

\section{SegFly Dataset}
\label{sec_segfly_dataset}

\subsection{Raw Data}
\label{subsec_raw_data}
We build our dataset upon the geo-referenced aerial imagery of the OccuFly benchmark~\cite{gross2025occufly} (CVPR 2026), containing 9 distinct scenes.
Data acquisition relies on two DJI UAV platforms: a Phantom~4 RTK (P4)~\cite{dji2025phantom4rtk} and a DJI Mavic~3 Enterprise Series (M3-ES)~\cite{dji2025mavic3enterprise}.
The P4 provides RGB imagery at a resolution of $5472\times3648$ pixels, whereas the M3-ES is the RGB–thermal-capable Mavic~3 Thermal configuration, providing $4000\times3000$ RGB frames and $640\times512$ long-wave infrared frames via software-based time-synchronization between both sensors.
OccuFly scenes 1, 2, and 6-8 flown with the P4 thus yield RGB-only imagery, while scenes 3-5 and 9 flown with M3T produce paired RGB–T views.
In total, our dataset leverages 20{,}606 geo-referenced RGB images and 15{,}007 geo-referenced thermal images with high multi-view overlap as the raw data source for 2D–3D–2D pseudo-label generation and RGB–T image registration.
Finally, following OccuFly~\cite{gross2025occufly}, we perform the 2D-to-3D label lifting per scene with OccuFly's total of 586 manually annotated RGB images captured at \SI{50}{m}, but additionally exploit geo-referenced RGB and thermal imagery from \SI{40}{m} and \SI{30}{m} flights for the 3D-to-2D pseudo-labeling stage, effectively amortizing the fixed reconstruction and annotation cost at one altitude for a substantially larger image set without additional manual labeling.

\subsection{Dataset Statistics}
\label{subsec_dataset_statistics}

\begin{figure}[t]
\centering
\renewcommand{\arraystretch}{0.9}
\begin{minipage}[b]{0.44\textwidth}
    \centering
    \includegraphics[width=\linewidth]{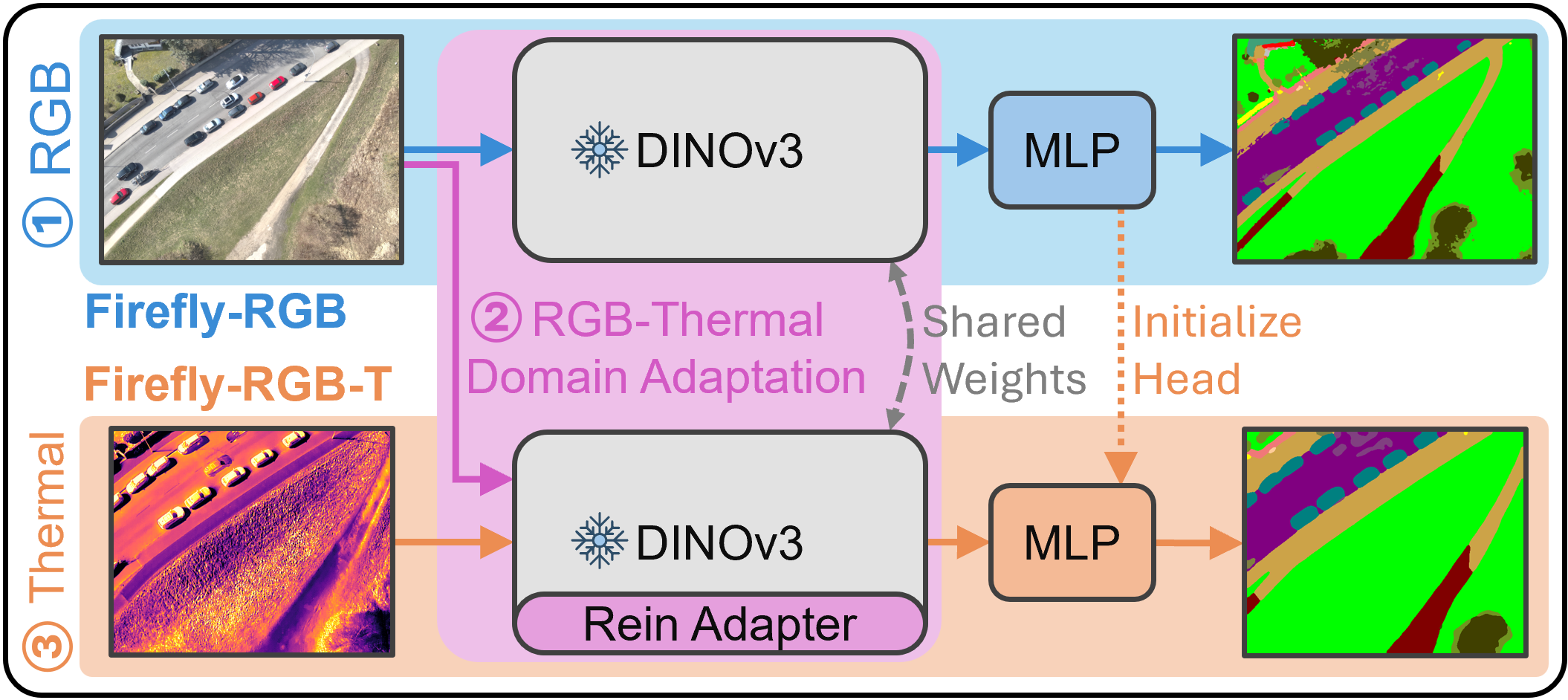}
    \caption{Firefly overview (see \cref{sec_firefly_method}).}
    \label{fig_firefly}
\end{minipage}
\hfill
\begin{minipage}[b]{0.55\textwidth}
    \centering
    \includegraphics[width=\linewidth]{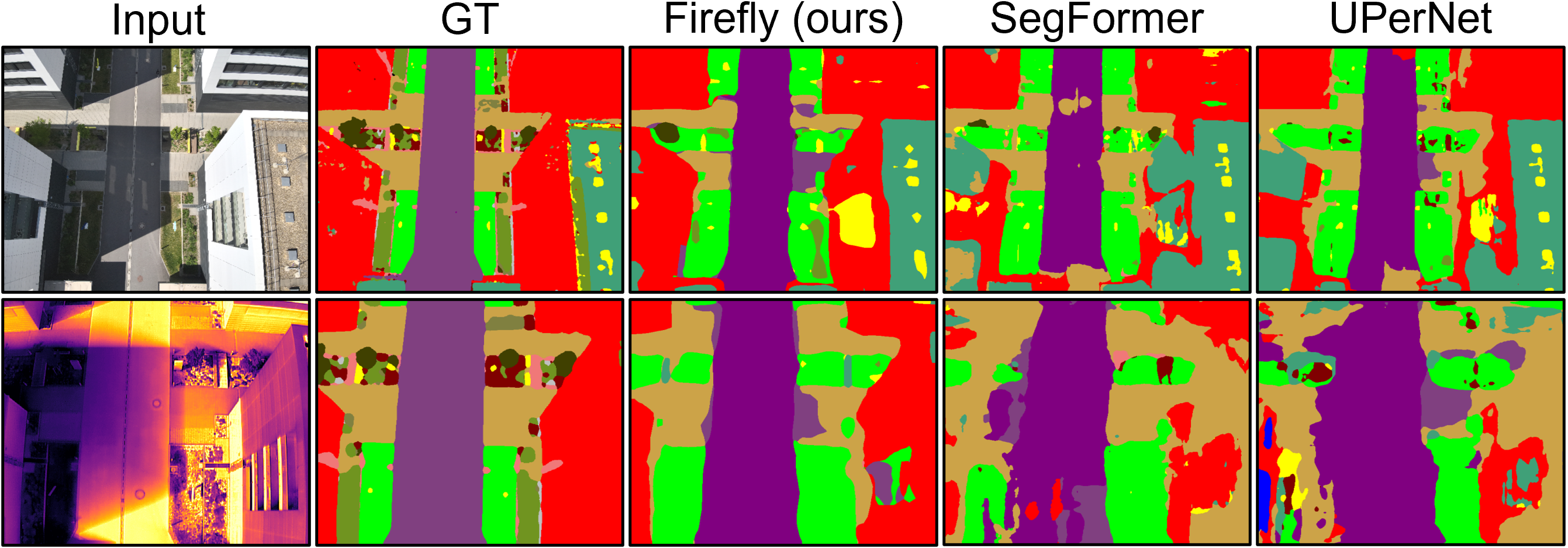}
    \caption{Qualitative segmentation results (\cref{sec_experiments}).}
    \label{fig_qualitative_results}
\end{minipage}
\renewcommand{\arraystretch}{1.0}
\end{figure}

Overall statistics are summarized in~\cref{tab_semantic_datasets}.
The dataset adheres to OccuFly's~\cite{gross2025occufly} data organization and comprises 9 scenes covering rural, urban, and industrial environments across all four seasons at \SI{50}{m}, \SI{40}{m}, and \SI{30}{m} altitude.
We present SegFly as an RGB-variant (\mbox{\emph{\textbf{SegFly-RGB}}}) from OccuFly scenes 1-9, and an RGB-T-variant (\mbox{\emph{\textbf{SegFly-RGB-T}}}) from OccuFly scenes 3, 4, 5, and 9. 
Importantly, we retain the original OccuFly~\cite{gross2025occufly} train/val/test split and augment it with additional RGB–T image pairs and 2D semantic annotations, effectively extending the underlying benchmark to a multi-modal 2D segmentation setting.
Note that this implies that SegFly-RGB-T shares identical val/test splits, an intentional design choice to ensure sufficient sample counts and preserve the original organizational structure of the OccuFly dataset.
Details regarding splits, seasons, environments, and altitude-wise sample counts are provided in the supplementary material.
In summary, the RGB modality contains 20{,}606 samples, while the thermal modality comprises 15{,}007 samples.
Every thermal sample forms an RGB–T pair, of which the RGB image was aligned to the thermal image by our registration pipeline described in \cref{subsec_rgbt_registration}.
Finally, since our pseudo-labeling pipeline builds on OccuFly’s classical SfM+MVS reconstruction under a static-scene assumption, dynamic objects or thin structures cannot always be reliably reconstructed in 3D.
To prevent these ambiguous classes from propagating into the final pseudo-labels and to ensure high-quality supervision, we apply a semantic post-processing step in which the rare classes \textit{Rock} (average class frequency \SI{0.063}{\%}) and \textit{CableTower} (\SI{0.002}{\%}) are merged into \textit{GroundObstacle}, and the classes \textit{Person}, \textit{Bicycle}, \textit{FlyingAnimals}, \textit{Cables}, and \textit{Cranes} (\SI{0.006}{\%} combined) are mapped to an unlabeled category.
Consequently, the final label set consists of 15 semantic classes, and $1.54/$\SI{1.78}{\%} of RGB/thermal pixels are unlabeled, primarily in images that cover the 3D scene boundaries during rendering. We retain these images to preserve realistic viewpoint statistics.

\begin{table}[t]
\caption{2D-3D-2D label transfer quality and efficiency (see \cref{subsec_dataset_evaluation}) by rendering a second set of manual GT that was \emph{not} included during data generation.}
\label{tab_segfly_accuracy_efficiency}
\centering
\renewcommand{\arraystretch}{1.0}
\resizebox{0.9\columnwidth}{!}{
\begin{tabular}{l|c|l|c|c|c|c}
\toprule
& & \multicolumn{4}{c|}{Altitude-wise Accuracy/FWmIoU [\%] $\uparrow$ } & [\%] of Manual  \\
\cline{3-6}
Dataset & Modality & \multicolumn{1}{c|}{50m} & 40m & 30m & combined & Annotation $\downarrow$  \\
\midrule
SegFly-RGB (Ours) & RGB & $91.30/85.90$ & $91.74/85.57$ & $91.38/86.10$ & $\mathbf{91.27/85.63}$ & $\mathbf{2.84}$ \\
\hline
SegFly-RGB-T (Ours) & Thermal & $84.93/75.70$ & $88.37/81.16$ & $89.64/82.88$ & $\mathbf{87.65/79.46}$ & $\mathbf{0.00}$ \\
\bottomrule
\end{tabular}
}
\end{table}

\subsection{Dataset Evaluation}
\label{subsec_dataset_evaluation}

\begin{figure}[t!]
	\centering
	\includegraphics[width=1.0\linewidth]{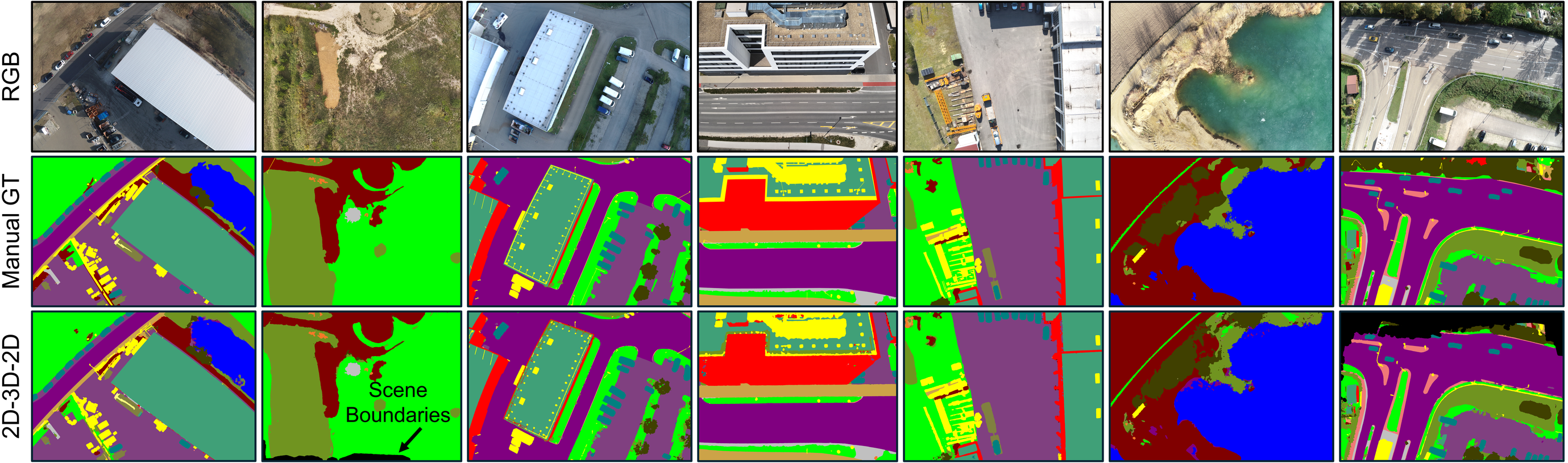}
	\caption{Comparison of 2D-3D-2D pseudo-labels vs. manual ground-truth (\cref{subsec_dataset_evaluation}).}
	\label{fig_pseudo_vs_manual}    
\end{figure}

\textbf{2D-3D-2D Label Transfer Accuracy and Efficiency.}\;
To ensure unbiased evaluation, we compare the generated pseudo-labels against \emph{additional} manually annotated ground-truth that was \emph{not} used during data generation (135 RGB images and 60 thermal images, distributed across all altitudes).
We summarize labeling accuracy and efficiency in~\cref{tab_segfly_accuracy_efficiency}.
Overall, the automatically generated RGB pseudo-labels achieve \SI{91}{\%} global pixel accuracy and \SI{86}{\%} class-frequency-weighted mean intersection over union (FWmIoU).
Notably, this performance is achieved even though only \SI{2.84}{\%} of the RGB data was manually annotated during data generation, while the remaining \SI{97.16}{\%} was generated automatically.
For thermal imagery, which is created entirely without manual annotation, we obtain $88/$\SI{79}{\%} accuracy/FWmIoU.
Furthermore, the proposed multi-altitude strategy substantially improves labeling efficiency: manual annotation and semantic reconstruction are performed at a single altitude, and the results are rendered into RGB and thermal images captured at multiple altitudes. 
Consequently, the fixed costs of reconstruction and manual labeling are amortized across a significantly larger number of images, thereby improving scalability without requiring additional human effort.
Finally, Fig. \ref{fig_pseudo_vs_manual} provides a qualitative comparison between manual ground truth and the generated pseudo-labels, highlighting the effectiveness of our label transfer. Despite the exceptionally high fidelity of OccuFly's~\cite{gross2025occufly} manual GT, the pseudo labels exhibit strong structural and semantic consistency, supporting the reliability of the proposed method.\\

\noindent\textbf{3D-to-2D Rendering Evaluation.}\;
We ablate every major step of our proposed point-based semantic rendering pipeline by rendering and comparing with manual semantic ground-truth that was used for data generation, effectively synthesizing the ground-truth (see \cref{tab_3D_2D_ablation}).
This ablation shows that our proposed configuration yields the best results.
We further compare our method with the rendering pipeline of UAVScenes~\cite{wang2025uavscenes} (ICCV 2025), which applies full manual 3D annotation followed by Screened Poisson Surface Reconstruction (SPSR).
Although SPSR reconstructs a watertight mesh, the UAVScenes rendering procedure produces blended colors that do not match explicit semantic labels, yielding \SI{64.07}{\%} unlabeled pixels and an annotation accuracy/FWmIoU of only $34.92/34.48\%$. 
To mitigate these semantically ambiguous pixels, UAVScenes applies a second round of manual refinement, which undermines efficient and scalable data generation.
In contrast, our pseudo-labels do not require manual refinement and achieve $90.08/85.14\%$ annotation accuracy/FWmIoU.
Taken together, these results validate our point-based semantic rendering pipeline and highlight its value for scalable, high-fidelity semantic data generation.\\

\noindent\textbf{Semantic Consistency between Manual and Pseudo-Labeled Data.}\;
We verify that pseudo-labeling does not alter the underlying semantic
distribution: per-class pixel frequencies computed from
manual annotations and from our 2D--3D--2D pseudo-labels differ by only
$1.06$\% on average. A detailed analysis and visualization of these distributions are provided in the supplementary material.\\

\noindent\textbf{RGB-T Registration Evaluation.}\;
To quantitatively validate the accuracy of the RGB-T registration, we utilize the second set of manual ground-truth that was \emph{not} used for data generation (60 pairs across all scenes and altitudes). Specifically, we apply the identical projection, lens-distortion transfer, and cropping procedure (used to construct $\mathbf{I}_n^{\text{RGB,reg}}$) to the RGB ground-truth of that set.
The resulting aligned labels are compared with the corresponding thermal labels, providing a direct measure of semantic-level registration accuracy. 
Our method yields a strong RGB-T registration accuracy/FWmIoU of $87.05/77.65\%$.
We further ablate our method by neglecting thermal lens distortion transfer, resulting in inferior performance of $84.08/73.63\%$.
Finally, we qualitatively compare our registration results with established RGB-T aligned datasets (illustrated in the supplementary), and show that SegFly-RGB-T exhibits stronger registration performance.
In summary, high quantitative performance and visual superiority confirm that our registration achieves precise alignment across both modalities.

\section{Benchmark Experiments}
\label{sec_experiments}

\begin{table}[t!]
\centering
\renewcommand{\arraystretch}{0.9}
\begin{minipage}[t]{0.49\textwidth}
    \centering
    \caption{RGB semantic segmentation results (see \cref{sec_experiments}) on SegFly-RGB test set.}
    \label{tab_rgb_results}
    \small
    \resizebox{0.8\linewidth}{!}{
    \begin{tabular}{l|cc|cc}
    \toprule
    & \multicolumn{2}{c|}{\textbf{Train Set}} & \multicolumn{2}{c}{\textbf{All Classes} [\%]} \\
    \cmidrule(lr){2-3} \cmidrule(lr){4-5}
    Method & \multicolumn{1}{c|}{Manual GT} & SegFly-RGB & Acc. & FWmIoU \\
    \midrule
    \rowcolor[HTML]{E6E6E6} 
    \multicolumn{5}{l}{\textbf{Eval on Manual GT}} \\
    \midrule
    \multirow{2}{*}{UPerNet~\cite{xiao2018upernet}} & \cmk & & $52.08$ & $52.46$ \\
                                                    & & \cmk & $\textbf{52.13}$ & $\textbf{54.36}$ \\
    \midrule
    \multirow{2}{*}{SegFormer~\cite{xie2021segformer}} & \cmk & & $50.05$ & $48.50$ \\
                                                       & & \cmk & $\textbf{51.48}$ & $\textbf{51.36}$ \\
    \midrule
    \multirow{2}{*}{Firefly (ours)} & \cmk & & $47.81$ & $45.81$ \\
                                    & & \cmk & $\textbf{53.40}$ & $\textbf{54.90}$ \\
    \midrule
    \rowcolor[HTML]{E6E6E6} 
    \multicolumn{5}{l}{\textbf{Eval on SegFly-RGB}} \\
    \midrule
    \multirow{2}{*}{UPerNet~\cite{xiao2018upernet}} & \cmk & & $47.72$ & $43.72$ \\
                                                    & & \cmk & $\textbf{48.38}$ & $\textbf{46.73}$ \\
    \midrule
    \multirow{2}{*}{SegFormer~\cite{xie2021segformer}} & \cmk & & $46.02$ & $41.74$ \\
                                                       & & \cmk & $\textbf{48.01}$ & $\textbf{44.46}$ \\
    \midrule
    \multirow{2}{*}{Firefly (ours)} & \cmk & & $45.72$ & $42.34$ \\
                                    & & \cmk & $\textbf{49.70}$ & $\textbf{50.44}$ \\
    \bottomrule
    \end{tabular}
    }
\end{minipage}
\hfill
\begin{minipage}[t]{0.49\textwidth}
    \centering
    \caption{RGB-T semantic segmentation results (\cref{sec_experiments}) on SegFly-RGBT test set.}
    \label{tab_rgbt_results}
    \small
    \resizebox{\linewidth}{!}{
    \begin{tabular}{l|cc}
    \toprule
    Method & \multicolumn{1}{c}{Acc.} & \multicolumn{1}{c}{FWmIoU} \\
    \midrule
    \rowcolor[HTML]{E6E6E6} 
    \multicolumn{3}{l}{\textbf{RGB-T Finetuning from Scratch}} \\
    \midrule
    UPerNet~\cite{xiao2018upernet}        & $31.36$ & $27.82$ \\
    SegFormer~\cite{xie2021segformer}     & $\underline{33.26}$ & $\underline{28.33}$ \\
    Firefly-RGB (ours)                        & $\textbf{37.48}$ & $\textbf{35.20}$ \\
    \midrule
    \rowcolor[HTML]{E6E6E6} 
    \multicolumn{3}{l}{\textbf{RGB Pretraining + RGB-T Finetuning}} \\
    \midrule
    UPerNet~\cite{xiao2018upernet}        & $33.48$ & $32.21$ \\
    SegFormer~\cite{xie2021segformer}     & \underline{$34.26$} & \underline{$30.66$}  \\
    Firefly-RGB (ours)                    & $\textbf{37.67}$ & $\textbf{37.38}$ \\
    \midrule
    \rowcolor[HTML]{E6E6E6} 
    \multicolumn{3}{l}{\textbf{RGB Pretraining $\rightarrow$ RGB-T Adaptation $\rightarrow$ RGB-T Finetuning}} \\
    \midrule
    UPerNet~\cite{xiao2018upernet} + Rein~\cite{wei2024rein}    &  $34.74$ & $31.98$ \\
    SegFormer~\cite{xie2021segformer} + Rein~\cite{wei2024rein} &  \underline{$35.75$} & \underline{$33.48$} \\
    Firefly-RGB-T (ours)                                        &  $\textbf{41.06}$ & $\textbf{43.83}$ \\
    \bottomrule
    \end{tabular}
    }
\end{minipage}
\end{table}
\renewcommand{\arraystretch}{1.0}

\noindent\textbf{RGB Semantic Segmentation.}\;
We benchmark SegFly-RGB for RGB semantic segmentation by training the established segmentation models UPerNet~\cite{xiao2018upernet} and SegFormer~\cite{cho2024catseg} with their original architectures and recommended optimization settings, and by comparing them to our Firefly-RGB baseline under the same training protocol.
We consider two supervision regimes, namely training on the limited manual ground-truth of OccuFly and training on the full SegFly-RGB pseudo-label set, and evaluate each model both on the manual ground-truth test split and on the SegFly-RGB test split.
Results in \cref{tab_rgb_results} and \cref{fig_qualitative_results} show that, when supervised with SegFly-RGB, Firefly-RGB achieves the highest accuracy and class-frequency-weighted mean intersection over union (FWmIoU) across both evaluation splits, outperforming UPerNet and SegFormer.
We provide class-wise metrics in the supplementary material.
More importantly, across all three architectures and both evaluation protocols, models trained on SegFly-RGB consistently outperform those trained only on manual ground-truth, despite the fact that SegFly-RGB pseudo-labels are generated automatically from a small fraction of human annotations.
This demonstrates that our 2D-3D-2D pseudo-label pipeline provides a more effective supervisory signal than sparse manual labels alone and enables higher accuracy without additional human effort.
Additionally, we ablate the MLP head of our Firefly baseline with ASPP~\cite{chen2018aspp} and DPT~\cite{ranftl2021dpt}.
Results in the supplementary material show that our choice of MLP offers a more favorable trade-off between accuracy and model complexity.\\

\noindent\textbf{Unsupervised RGB Semantic Segmentation.}\;
To verify that the improvements in \cref{tab_rgb_results} are not merely due to training on more RGB images, we perform an additional control experiment. Recall that the manual supervision regime uses only the $586$ manually annotated SegFly-RGB frames ($\approx 3\%$ of the dataset), whereas the pseudo-label regime leverages all $20{,}606$ SegFly-RGB images. In both cases, performance is evaluated on the same held-out test scenes, using either the manual annotations or the corresponding SegFly-RGB pseudo-labels. As a control, we train PriMaPs~\cite{han2024primaps}, a recent unsupervised segmentation method, on the same full set of SegFly-RGB images but without labels, thereby matching the number of training images while removing the supervisory signal from our pseudo-labels. PriMaPs achieves $31.19/16.84\%$ and $31.64/15.31\%$ Acc/FWmIoU on the manually annotated test set and the full SegFly-RGB test split, respectively. In contrast, our Firefly-RGB method improves these scores by $+71/+226\%$ and $+57/+229\%$ Acc/FWmIoU, respectively. These results demonstrate that the gains in \cref{tab_rgb_results} stem from the additional supervision provided by our 2D--3D--2D pseudo-label framework rather than a larger sample count.\\

\begin{table}[t!]
\centering
\renewcommand{\arraystretch}{1.0}
\begin{minipage}[t]{0.49\textwidth}
    \centering
    \caption{Ablation on the 3D-to-2D pseudo-labels from our proposed point-based semantic rendering pipeline against manual ground-truth of test set scene 8 (\cref{subsec_dataset_evaluation}).}
    \label{tab_3D_2D_ablation}
    \small
    \renewcommand{\arraystretch}{1.0}
    \resizebox{1.0\linewidth}{!}{
    \begin{tabular}{l|ccc|cc}
    \toprule
     & Occlusion & Splatting based & Depth Guided & Accuracy & FWmIoU \\
    Method & Filtering & Densification & Label Distribution & [\%] & [\%] \\
    \midrule
    (A)        &        & \cmk & \cmk &  $89.78$ & $84.58$      \\
    (B)        & \cmk   &      & \cmk & $89.54$ & $84.35$       \\
    (C)        & \cmk   & \cmk &      & $83.95$ & $80.49$        \\
    \hline
    (D) (ours) & \cmk   & \cmk & \cmk & $\mathbf{90.08}$ & $\mathbf{85.14}$       \\
    \bottomrule
    \end{tabular}
    }
\end{minipage}
\hfill
\begin{minipage}[t]{0.49\textwidth}
    \centering
    \caption{Evaluation on SegFly's capability to improve Vision Foundation Models for aerial RGB and RGB-T semantic segmentation, by zero-shot evaluation and fine-tuning on SegFly (see \cref{sec_experiments}).}
    \label{tab_vision_foundation_models}
    \small
    \resizebox{1.0\linewidth}{!}{
    \begin{tabular}{c|c|cc|c}
    \toprule
    Method & \multicolumn{1}{c|}{Eval Type}  & \multicolumn{1}{c|}{Eval Set} & Acc. & FWmIoU  \\
    \midrule
    \rowcolor[HTML]{E6E6E6} 
    \multicolumn{5}{l}{\textbf{SegFly-RGB Semantic Segmentation}} \\
    \midrule
    \multirow{1}{*}{CatSeg~\cite{cho2024catseg} } &  Zero-Shot  & \multicolumn{1}{c|}{CART~\cite{lee2024cart}}    &    \multicolumn{1}{c|}{$4.32$}   &  \multicolumn{1}{c}{$15.03$}   \\
    \cline{2-5}
    (CVPR 2024)  &  Finetuned SegFly-RGB & \multicolumn{1}{c|}{CART~\cite{lee2024cart}}  &   $\mathbf{11.35}$  &  $\mathbf{17.45}$     \\
    \midrule
    \rowcolor[HTML]{E6E6E6} 
    \multicolumn{5}{l}{\textbf{SegFly-RGB-T Semantic Segmentation}} \\
    \midrule
    \multirow{1}{*}{AnyThermal~\cite{mahes2026anythermal}} & Zero-Shot & \multicolumn{1}{c|}{Kust4K~\cite{ouyang2025kust4k}}     &   $46.02$    &  $35.06$    \\
    \cline{2-5}
    (ICRA 2026)    & Finetuned SegFly-RGB-T   &   \multicolumn{1}{c|}{Kust4K~\cite{ouyang2025kust4k}} &  $\mathbf{58.73}$   & $\mathbf{51.75}$   \\
    \bottomrule
    \end{tabular}
    }
\end{minipage}
\end{table}
\noindent\textbf{Thermal Semantic Segmentation.}\;
We ablate Firefly's domain adaptation module Rein~\cite{wei2024rein} with LoRA~\cite{hu2022lora} and Adapter~\cite{chen2022adapter}.
Detailed results presented in the supplementary show that our choices of MLP and Rein yield superior results.
Moreover, we benchmark thermal semantic segmentation on SegFly-RGB-T by extending the three-stage Firefly training scheme (discussed in \cref{sec_firefly_method}) to UPerNet~\cite{xiao2018upernet} and SegFormer~\cite{xie2021segformer}.
Results in \cref{tab_rgbt_results} and \cref{fig_qualitative_results} highlight that for all models, RGB pretraining consistently improves thermal performance over thermal-only training, and adding cross-modal adaptation further boosts accuracy and FWmIoU for SegFormer and especially for Firefly, where Firefly-RGB-T achieves the best results with $41.06\%$ accuracy and $43.83\%$ FWmIoU.
These findings show that SegFly-RGB and SegFly-RGB-T, together with the proposed RGB pretraining and RGB-T adaptation scheme, provide an effective setting for learning thermal semantic segmentation and for transferring knowledge from RGB to the thermal modality, most prominently for our Firefly baseline.
We provide additional qualitative examples and full class-wise metrics in the supplementary.\\

\noindent\textbf{Semantic Segmentation with RGB-Thermal Fusion.}\;
To assess the effect of RGB–thermal fusion, we extend our Firefly baseline to a multimodal variant on SegFly-RGB-T by concatenating RGB and thermal encoder features before the segmentation head and training under the same protocol as the thermal-only model.
This fusion improves Firefly to $42.33/44.26\%$ Acc/FWmIoU, corresponding to a $+1.27/+0.43\%$ gain over the thermal-only variant.
These results suggest that RGB provides only marginal additional information beyond thermal under our training protocol, highlighting the strength of our RGB→thermal adaptation in effectively distilling RGB supervision into the thermal modality.\\

\noindent\textbf{Vision Foundation Models.}\;
We evaluate how SegFly supports state-of-the-art vision foundation models for aerial RGB and thermal segmentation, presented in ~\cref{tab_vision_foundation_models}.
We consider CatSeg~\cite{cho2024catseg} (CVPR 2024) and AnyThermal~\cite{mahes2026anythermal} (ICRA 2026) for RGB and thermal semantic segmentation, respectively.
Following a cross-dataset evaluation, we first evaluate these pretrained models directly on an independent target dataset in a zero-shot manner. Subsequently, we fine-tune the models on SegFly and evaluate them again on the same target dataset. Keeping the evaluation set fixed allows us to quantify how fine-tuning on our data affects cross-dataset generalization.
On SegFly-RGB, CatSeg improves from $4.32/15.03\%$ accuracy/FWmIoU in zero-shot mode to $11.35/17.45\%$ accuracy/FWmIoU after fine-tuning.
On SegFly-RGB-T, AnyThermal improves from $46.02/35.06\%$ in zero-shot mode to $58.73/51.75\%$ accuracy/FWmIoU after fine-tuning.
Together, these gains highlight that our dataset provides effective supervision that generalizes beyond the training domain, significantly improving cross-dataset generalization.

\section{Conclusion}
\label{sec_conclusion}
We present SegFly as an aerial RGB-T benchmark built with a 2D-3D-2D paradigm that turns a small set of manual RGB annotations into dense RGB and registered RGB-T labels across multiple altitudes.
Experiments show that SegFly significantly improves RGB and thermal semantic segmentation for established architectures, recent aerial vision foundation models, and our proposed Firefly baseline.
Future work will address remaining limitations, including dynamic objects by 4D methods~\cite{song2025coda}, and the current restriction to daytime thermal imagery by investigating adversarial training strategies for robust adaptation across varying times of day~\cite{vertens2020heatnet,lee2024cart}.

\bibliographystyle{splncs04}
\bibliography{main}

\clearpage
\appendix
\thispagestyle{empty}   

\begin{center}
{\Large\bfseries SegFly: A Dataset and 2D-3D-2D Paradigm\\for Aerial RGB-Thermal Semantic Segmentation\\at Scale}\\[8pt]
    {\large Supplementary Material}\\[6pt]
\end{center}

\section{SegFly Dataset Statistics}
\begin{figure}[h!]
  \centering
  \begin{minipage}[t]{0.49\linewidth}
    \centering
    \includegraphics[width=\linewidth]{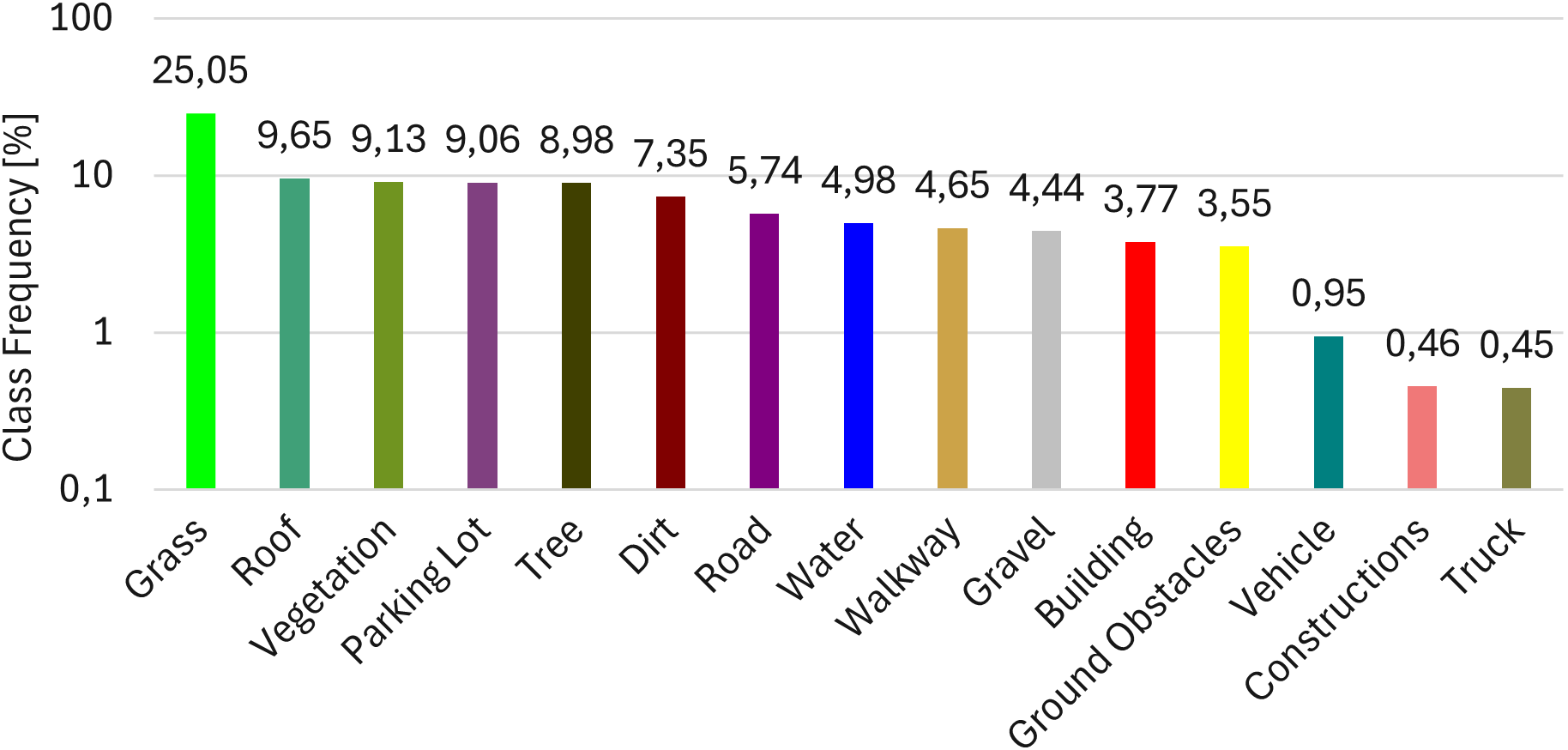}
    \captionof{figure}{SegFly-RGB class frequencies.}
    \label{fig_segfly_rgb}
  \end{minipage}
  \hfill
  \begin{minipage}[t]{0.49\linewidth}
    \centering
    \includegraphics[width=\linewidth]{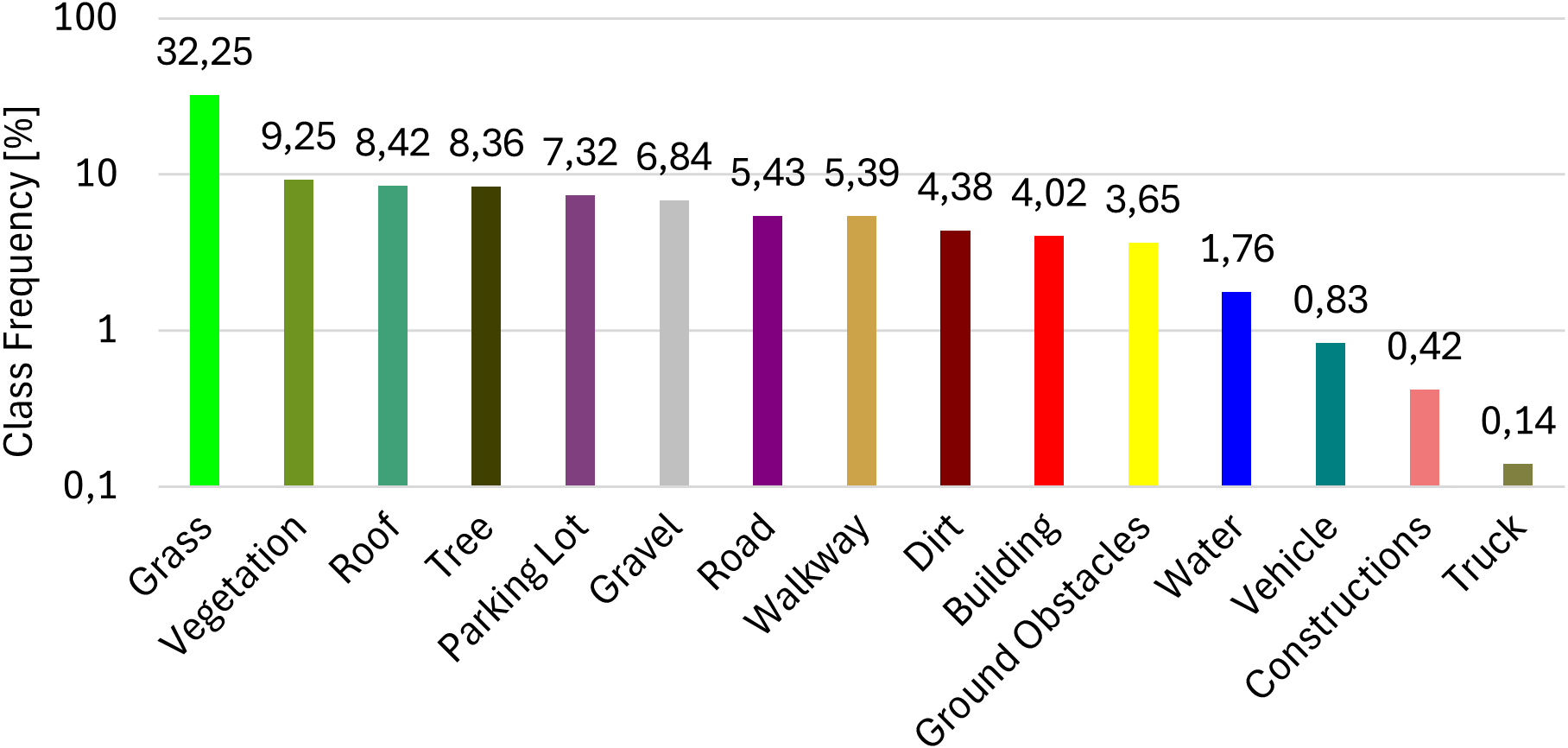}
    \captionof{figure}{SegFly-RGB-T class frequencies.}
    \label{fig_segfly_rgbt}
  \end{minipage}
\end{figure}

In addition to the SegFly dataset statistics discussed in \cref{sec_segfly_dataset} and summarized in \cref{tab_semantic_datasets}, we provide further details on seasons, environments, and altitude-wise sample counts for \mbox{SegFly-RGB} in \cref{tab_segfly_rgb}, and for \mbox{SegFly-RGB-T} in \cref{tab_segfly_rgbt}. Additionally, \cref{fig_segfly_rgb} and \cref{fig_segfly_rgbt} illustrate semantic class frequencies.

\begin{table}[h!]
\centering
\renewcommand{\arraystretch}{0.88}
\begin{minipage}[t]{0.49\textwidth}
    \centering
    \caption{SegFly-RGB dataset details.}
    \label{tab_segfly_rgb}
    \resizebox{0.9\linewidth}{!}{%
    \begin{tabular}{l|c|c|ccc|c}
    \toprule
    \multirow{2}{*}{Scene} & \multirow{2}{*}{Season} & \multirow{2}{*}{Environment} & \multicolumn{4}{c}{Number of Samples} \\
    \cline{4-7}
    & & & \SI{30}{m} & \SI{40}{m} & \SI{50}{m} & Total \\
    \midrule
    \rowcolor[HTML]{E6E6E6}
    \multicolumn{1}{l}{\textbf{Training}} & \multicolumn{5}{c}{} & $\mathbf{14,799}$ \\
    01       & Winter & Rural       & 512  & 469  & 406  & 1,387 \\
    02        & Winter & Urban       & 132  & 468  & 277  & 877  \\
    03      & Spring & Urban       & 1,427 & 1,315 & 871  & 3,613 \\
    04 & Spring & Industrial  & 1,565 & 1,258 & 1,140 & 3,963 \\
    05     & Summer & Rural       & 2,270 & 1,683 & 1,006 & 4,959 \\
    \midrule
    \rowcolor[HTML]{E6E6E6}
    \multicolumn{1}{c}{\textbf{Validation}} & \multicolumn{5}{c}{} & $\mathbf{1,965}$ \\
    06       & Spring & Urban       & 327  & 266  & 366  & 959  \\
    07     & Spring & Industrial  & 343  & 384  & 279  & 1,006 \\
    \midrule
    \rowcolor[HTML]{E6E6E6}
    \multicolumn{1}{l}{\textbf{Test}} & \multicolumn{5}{c}{} & $\mathbf{3,842}$ \\
    08   & Fall   & Industrial  & 383  & 316  & 183  & 882  \\
    09        & Spring & Urban       & 240  & 1,416 & 1,304 & 2,960 \\
    \midrule
    \textbf{Total} & \multicolumn{1}{c|}{} & \multicolumn{1}{c|}{} 
                   & $\mathbf{7,199}$ & $\mathbf{7,575}$ & $\mathbf{5,832}$ & $\mathbf{20,606}$ \\
    \bottomrule
    \end{tabular}
    }
\end{minipage}
\hfill
\begin{minipage}[t]{0.49\textwidth}
    \centering
    \caption{SegFly-RGB-T dataset details.}
    \label{tab_segfly_rgbt}
    \resizebox{0.99\linewidth}{!}{%
    \begin{tabular}{l|c|c|ccc|c}
    \toprule
    \multirow{2}{*}{Scene} & \multirow{2}{*}{Season} & \multirow{2}{*}{Environment} & \multicolumn{4}{c}{Number of Samples} \\
    \cline{4-7}
    & & & \SI{30}{m} & \SI{40}{m} & \SI{50}{m} & Total \\
    \midrule
    \rowcolor[HTML]{E6E6E6}
    \multicolumn{1}{l}{\textbf{Training}} & \multicolumn{5}{c}{} & $\mathbf{12,063}$ \\
    03      & Spring & Urban      & 1,350 & 1,199 & 824   & 3,373 \\
    04 & Spring & Industrial & 1,537 & 1,254 & 1,140 & 3,931 \\
    05   & Summer & Rural      & 2,163 & 1,623 & 973   & 4,759 \\
    \midrule
    \rowcolor[HTML]{E6E6E6}
    \multicolumn{1}{l}{\textbf{Validation / Test}} & \multicolumn{5}{c}{} & $\mathbf{2,944}$ \\
    09        & Spring & Urban      & 236   & 1,404 & 1,304 & 2,944 \\
    \midrule
    \textbf{Total} & \multicolumn{1}{c|}{} & \multicolumn{1}{c|}{} 
                   & $\mathbf{5,286}$ & $\mathbf{5,480}$ & $\mathbf{4,241}$ & $\mathbf{15,007}$ \\
    \bottomrule
    \end{tabular}
    }
\end{minipage}
\end{table}
\renewcommand{\arraystretch}{1.0}

\vspace{-1cm}

\section{Semantic Consistency between Manual and Pseudo-Labeled Data}

\begin{figure}[h!]
	\centering
	\includegraphics[width=0.75\linewidth]{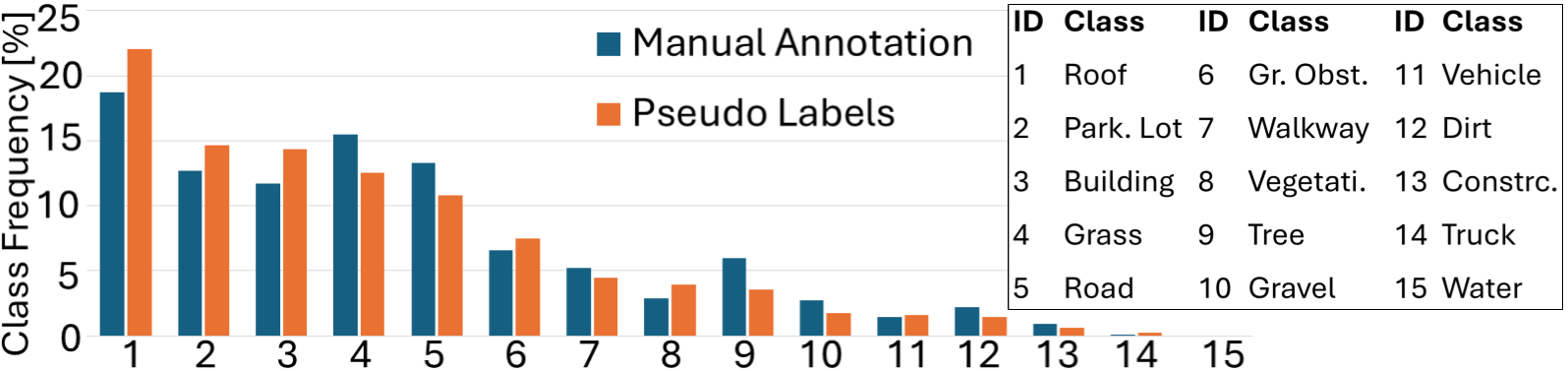}
	\caption{Semantic Consistency between manual ground-truth and SegFly pseudo-labels.}
	\label{fig_domain_shift}    
\end{figure}

A natural question arising from the use of pseudo-labels is whether replacing manual annotations with pseudo-labels alters the underlying data distribution seen by segmentation models.
To assess this, we compare the semantic class-frequency distributions induced by manual ground-truth  our 2D--3D--2D pseudo-labels.
As shown in \cref{fig_domain_shift}, across all classes, the average discrepancy is only $1.06$\%, and no single class exhibits a substantial deviation.
This analysis indicates that our pseudo-labeled dataset preserves the semantic composition of the manually annotated data and does not introduce a measurable domain shift in class-occurrence statistics.

\section{Effect of Image Resolution during Data Generation}
We study the impact of input resolution on pseudo-label quality by re-running the full 2D-3D-2D pipeline on SegFly test scene~9, in which all RGB frames are downsampled to one quarter of their original resolution (i.e., both height and width are reduced by a factor of two) prior to reconstruction and label transfer. Using the held-out manual ground-truth, the resulting pseudo-labels achieve $86.22$\% global pixel accuracy and $77.42$\% FWmIoU, slightly exceeding the performance of our default full-resolution configuration on the same scene. These results indicate that the proposed geometry-driven pipeline is robust to moderate reductions in input resolution and can even benefit from the denoising effect of downsampling, without incurring a loss in semantic annotation quality.

\section{RGB-T Image Registration}
\begin{figure}[h!]
	\centering
	\includegraphics[width=.99\linewidth]{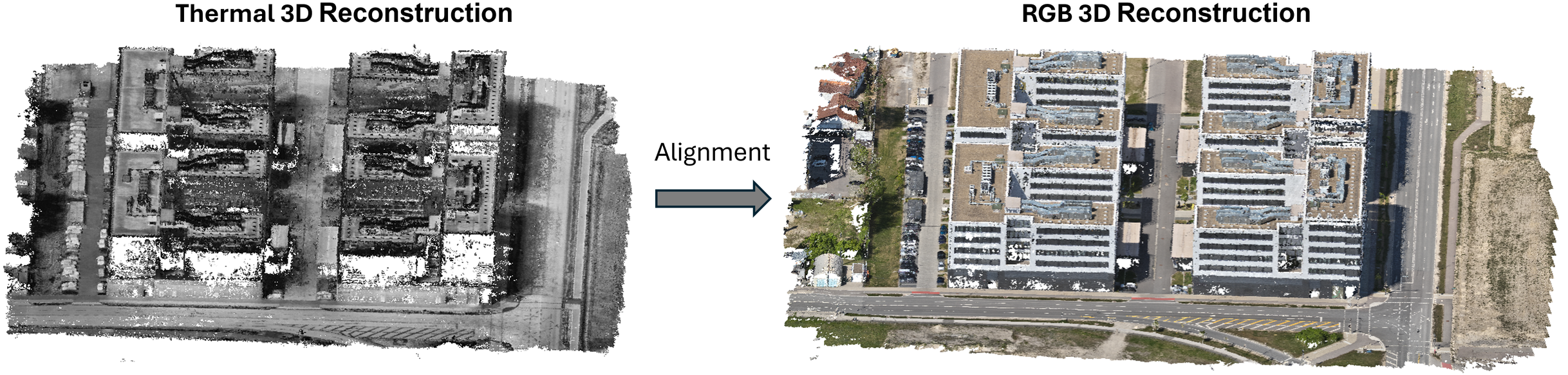}
	\caption{Alignment of thermal and RGB 3D reconstructions used for RGB-T registration.}
	\label{fig_rgb_thermal_alignment}    
\end{figure}

\begin{figure}[h!]
	\centering
	\includegraphics[width=.99\linewidth]{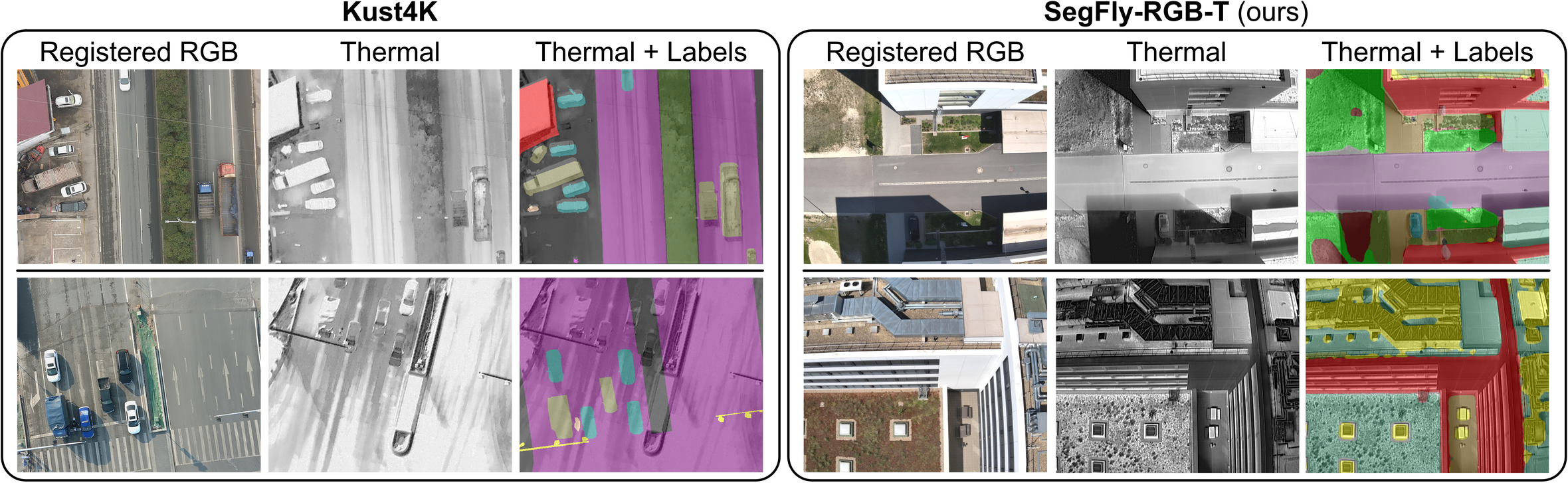}
	\caption{Comparison of RGB-T image registration results of Kust4K~\cite{ouyang2025kust4k} and SegFly-RGB-T (ours).}
	\label{fig_kust4k_vs_segfly}    
\end{figure}

In addition to the discussion and quantitative metrics reported in \cref{subsec_dataset_evaluation}, we further visualize and qualitatively evaluate the RGB-T image registration results of our proposed method (\cref{subsec_rgbt_registration}).
Specifically, \cref{fig_rgb_thermal_alignment} shows the reconstructed 3D point clouds of both the RGB and thermal modalities, which are at the core of our proposed registration method.
Moreover, \cref{fig_kust4k_vs_segfly} compares our dataset with the established RGB-T-aligned Kust4K dataset~\cite{ouyang2025kust4k}.
In SegFly, RGB and thermal views exhibit strong alignment, and semantic labels projected from RGB closely match thermal structures. 
In Kust4K, overlays of the labels on thermal images instead reveal a systematic translation of the labels, roughly aligned with the UAV flight direction, and some sequences with strong yaw exhibit severe local misalignment between modalities.
We attribute these differences to the underlying registration strategies:
Kust4K performs pairwise 2D registration using SIFT~\cite{lowe2004sift} feature matching with additional manual adjustment. 
On a moving platform with six degrees of freedom motion and strong depth variation, a single 2D transform can only approximate this geometry, so residual parallax and pose errors appear.
Our 2D–3D–2D framework instead reconstructs RGB and thermal imagery in 3D, aligns the point clouds, and renders semantic labels using calibrated intrinsics and lens distortion transfer.
This geometry-based alignment enforces multi-view consistency across the scene and accounts for parallax and varying pose differences, resulting in visually superior alignment.

\vspace{-1cm}

\section{Quality-Efficiency-Tradeoff during Data Generation}

To study the trade-off between manual annotation effort and the resulting pseudo-labeling quality, we conduct an ablation by varying the fraction of RGB frames that are manually annotated and lifted to 3D.
Starting from our default setting with \SI{3.1}{\%} manual annotations, we reduce the supervision to \SI{2.1}{\%} and \SI{1.1}{\%} by sub-sampling the set of source views while keeping the reconstruction and registration pipeline fixed.
Results presented in \cref{tab_quality_efficiency_tradeoff} show that reducing the manual ratio from \SI{3.1}{\%} to \SI{2.1}{\%} causes a relatively minor decrease in semantic alignment between the resulting pseudo-labels and the manual ground-truth, whereas a further reduction to \SI{1.1}{\%} leads to a substantial degradation in quantitative metrics.
Note that we do not explore annotation ratios above \SI{3.1}{\%}, since the supervision is limited by available manual labels from OccuFly~\cite{gross2025occufly}.
Overall, these observations indicate that our geometry-driven registration exhibits a useful tolerance range for manual supervision, in which annotation effort could be reduced without a substantial loss in annotation quality, whereas more aggressive reductions result in degraded annotation performance.

\begin{table}[h!]
\caption{Quality-efficiency-tradeoff for reduced fractions of manually annotated images on scene 9 of the SegFly test set.}
\label{tab_quality_efficiency_tradeoff}
\centering
\renewcommand{\arraystretch}{1.0}
\resizebox{0.4\columnwidth}{!}{
\begin{tabular}{l|c|cc}
\toprule
\multicolumn{1}{c|}{Manual} & & \multicolumn{2}{c}{Pseudo-Labeling [\%] $\uparrow$ }  \\
\cline{3-4}
\multicolumn{1}{c|}{Annotation} & Modality & Accuracy & FWmIoU   \\
\midrule
\SI{1.1}{\%}  & RGB & $73.96$ & $62.49$ \\
\SI{0.0}{\%} & Thermal & $65.09$ & $53.19$ \\
\midrule
\SI{2.1}{\%}  & RGB & $86.36$ & $78.74$ \\
\SI{0.0}{\%} & Thermal & $78.20$ & $67.23$ \\
\midrule
\SI{3.1}{\%} (ours)  & RGB & $\mathbf{87.53}$ & $\mathbf{80.34}$ \\
\SI{0.0}{\%} (ours) & Thermal & $\mathbf{84.46}$ & $\mathbf{75.39}$ \\
\bottomrule
\end{tabular}
}
\end{table}

\section{Firefly Baseline}

\subsection{Firefly Ablation Details}

For the segmentation head, we ablate ASPP~\cite{chen2018aspp}, DPT~\cite{ranftl2021dpt}, and MLP~\cite{rumelhart1986mlp} on SegFly-RGB.
Detailed results in \cref{tab_head_ablations} show that both the ASPP and the MLP head outperform the DPT head, confirming that simple decoder structures are sufficient when combined with a strong encoder.
Between ASPP and the MLP, we observe similar performance, while the MLP uses substantially fewer parameters. We therefore adopt the MLP head, as it offers a more favorable trade-off between segmentation performance and model complexity.

\begin{table}[h!]
\centering
\caption{Ablation on different heads for Firefly on SegFly-RGB test set (\cref{sec_experiments}).}
\label{tab_head_ablations}
\small
\renewcommand{\arraystretch}{1.2}
\resizebox{\linewidth}{!}{
\begin{tabular}{l | c | cc | cccccccccccccccc}
\toprule
& Head & \multicolumn{2}{c|}{All Classes [\%]} & \multicolumn{15}{c}{Class-wise Intersection over Union (IoU) [\%]} \\
\cline{3-19}
Method & Params
& Acc. & FWmIoU & Road & Walkway & Dirt & Gravel & Grass & Vegetation & Tree & Grnd Obst. & Vehicle & Water & Building & Roof & Park. Lot & Constr. & Truck \\ 
\midrule
Firefly with ASPP~\cite{chen2018aspp} & $6.22$ M                                & $\mathbf{50.82}$ & $\underline{49.93}$ & $\underline{45.50}$ & $\underline{48.31}$ & $\mathbf{09.28}$ & $20.09$ & $\underline{47.60}$ & $05.76$ & $\mathbf{54.60}$ & $\mathbf{61.79}$ & $\mathbf{59.11}$ & - & $73.28$ & \underline{$55.17$} & $33.23$ & $\mathbf{06.24}$ & $00.78$ \\
Firefly with DPT~\cite{ranftl2021dpt} & $\underline{5.49}$ M                                     & $49.08$ & $49.63$ & $\mathbf{46.71}$ & $45.72$ & $05.88$ & $\mathbf{27.35}$ & $\mathbf{48.49}$ & $\underline{07.72}$ & $43.72$ & $56.71$ & \underline{$58.65$} & - & \underline{$73.56$} & $54.04$ & $\mathbf{36.44}$ & $02.50$ & $\mathbf{03.90}$ & \\
Firefly with MLP~\cite{rumelhart1986mlp} (ours)  & $\mathbf{2.36}$ M                       & $\underline{49.70}$ & $\mathbf{50.44}$ & $44.79$ & $\mathbf{51.85}$ & $\underline{07.43}$ & $\underline{24.19}$ & $47.14$ & $\mathbf{07.87}$ & $\underline{49.21}$ & \underline{$59.84$} & $56.48$ & - & $\mathbf{74.89}$ & $\mathbf{56.47}$ & \underline{$34.19$} & \underline{$02.95$} & \underline{$01.02$} & \\
\bottomrule
\end{tabular}
}
\end{table}

\noindent For parameter-efficient RGB–T domain adaptation, we compare LoRA~\cite{hu2022lora}, Adapter~\cite{chen2022adapter}, and Rein~\cite{wei2024rein} on SegFly-RGB-T.
As shown in \cref{tab_adapter_ablations}, the Adapter variant achieves the lowest overall segmentation performance.
LoRA and Rein perform roughly on par in terms of global pixel accuracy, but Rein attains a higher FWmIoU.
Since the adapter represents only a small fraction of the total model capacity compared to the frozen backbone, we adopt Rein as our default adapter, offering a favorable trade-off between parameter efficiency and segmentation performance.

\begin{table}[h!]
\centering
\caption{Ablation on adapters for Firefly on SegFly-RGB-T test set (\cref{sec_experiments}).}
\label{tab_adapter_ablations}
\small
\renewcommand{\arraystretch}{1.2}
\resizebox{\linewidth}{!}{
\begin{tabular}{l | c | cc | cccccccccccccccc}
\toprule
& Adapter & \multicolumn{2}{c|}{All Classes [\%]} & \multicolumn{15}{c}{Class-wise Intersection over Union (IoU) [\%]} \\
\cline{3-19}
Method & Params
& Acc. & FWmIoU & Road & Walkway & Dirt & Gravel & Grass & Vegetation & Tree & Grnd Obst. & Vehicle & Water & Building & Roof & Park. Lot & Constr. & Truck \\ 
\midrule
Firefly with LoRA~\cite{hu2022lora} & $\textbf{1.03}$ M                  & $\mathbf{41.33}$ & \underline{$42.10$} & $\mathbf{50.35}$ & $47.02$ & $\mathbf{08.42}$ & $\mathbf{05.51}$ & $32.11$ & $00.34$ & \underline{$15.90$} & $\mathbf{55.06}$ & \underline{$57.63$} & - & $56.89$ & \underline{$47.76$} & \underline{$12.65$} & $\mathbf{00.80}$ & $00.00$ \\
Firefly with Adapter~\cite{chen2022adapter} & $\underline{1.19}$ M              & $40.14$ & $41.42$ & \underline{$47.82$} & \underline{$47.41$} & $02.28$ & $03.67$ & $\mathbf{33.32}$ & \underline{$00.40$} & $15.71$ & $54.44$ & $\mathbf{59.38}$ & - & \underline{$57.12$} & $46.44$ & $11.70$ & \underline{$00.60$} & $\mathbf{00.05}$ \\
Firefly with Rein~\cite{wei2024rein} (ours) & $2.50$ M                 & \underline{$41.06$} & $\mathbf{43.83}$ & $46.69$ & $\mathbf{51.08}$ & \underline{$07.11$} & \underline{$05.16$} & \underline{$32.42$} & $\mathbf{00.85}$ & $\mathbf{17.20}$ & \underline{$54.90$} & $56.92$ & - & $\mathbf{62.69}$ & $\mathbf{49.72}$ & $\mathbf{14.72}$ & $00.51$ & $00.00$ \\
\bottomrule
\end{tabular}
}
\end{table}

\subsection{Firefly Method Details}

In addition to \cref{sec_firefly_method}, we provide technical details of our Firefly baseline (see \cref{fig_firefly}), a simple yet efficient method that addresses (1) RGB semantic segmentation and (2) thermal semantic segmentation via parameter-efficient fine-tuning with RGB-T domain adaptation.
Firefly training proceeds in three stages: (i) RGB pre-training, (ii) RGB-Thermal domain adaptation, and (iii) thermal fine-tuning.

\textbf{(i) RGB Pre-Training.}\;
In the first stage, we train the MLP head for semantic segmentation while keeping the DINOv3~\cite{simeoni2025dinov3} encoder fixed and without any adapters.
The model receives labeled RGB images and optimizes only the MLP parameters with a Dice loss~\cite{Milletari_2016_diceloss_vnet} on the segmentation masks.
This stage produces a strong RGB segmentation model that serves as the source for subsequent cross-modal adaptation.
Note that this stage constitutes standalone RGB-based semantic segmentation, but in the context of the full pipeline, we refer to it as RGB pre-training, since the head parameters initialize the thermal head in stage~(iii).

\textbf{(ii) RGB-Thermal Domain Adaptation.}\;
The second stage aligns thermal features to RGB features via self-supervised parameter-efficient domain adaptation.
We assume access to registered RGB and thermal image pairs.
For each pair, the RGB image is processed by the frozen DINOv3 encoder without adapters, yielding a sequence of RGB features at each transformer layer.
The corresponding thermal image is processed by the same encoder augmented with Rein adapters, which yields thermal features at the same layers.
For every layer, we encourage the thermal features to match the RGB features by minimizing a combination of mean squared error and Kullback-Leibler divergence on the feature distributions.
These losses are applied at all transformer layers without additional aggregation, and no segmentation labels are used in this stage.
Gradients are restricted to the Rein adapters, while the encoder remains fixed.
As a result, the adapters learn a thermal-specific residual that maps thermal inputs into the RGB feature space while leaving the RGB branch unchanged.

\textbf{(iii) Thermal Fine-Tuning.}\;
After domain adaptation, we introduce the dedicated thermal head.
Its weights are initialized as an exact copy of the RGB head obtained in stage one.
We then fine-tune on labeled thermal images while freezing the encoder and the RGB head and updating the thermal head together with the Rein adapters.
The objective is again the Dice loss on thermal segmentation masks.
This stage refines the thermal head and the thermal adapters for the target modality, starting from feature representations that have already been aligned to RGB by the second stage.

In summary, Firefly keeps the DINOv3 encoder fixed and restricts modality-specific learning to lightweight MLP heads and a small number of thermal Rein adapters.
With this approach, RGB pre-training establishes a strong source model, RGB-Thermal domain adaptation aligns thermal features to the RGB representation via unsupervised feature matching, and thermal fine-tuning specializes the thermal head and adapters for accurate thermal semantic segmentation while preserving the original performance on the RGB modality.

\section{Benchmark Experiments}

In addition to the discussion of our experiments regarding RGB semantic segmentation (\cref{tab_rgb_results}), thermal semantic segmentation (\cref{tab_rgbt_results}), and vision foundation models (\cref{tab_vision_foundation_models}), we provide class-wise results for the same experiments in \cref{tab_rgb_results_supp,tab_rgbt_results_supp,tab_vision_foundation_models_supp}, respectively.
Overall, the class-wise results are consistent with the findings of the aggregated results discussed in the manuscript.
Notably, dominant, large-footprint classes (\eg, road, walkway, grass, building, roof) achieve consistently strong IoUs across models and training regimes, whereas rare or small-footprint categories (\eg, truck and construction) exhibit relatively low and unstable IoUs for all methods.
We attribute the latter primarily to class imbalances rather than method-specific failures.
For individual class frequencies, refer to \cref{fig_segfly_rgb,fig_segfly_rgbt}.

\vspace{-0.5cm}
\begin{table}[h!]
\centering
\caption{RGB semantic segmentation results on SegFly-RGB test set.}
\label{tab_rgb_results_supp}
\small
\renewcommand{\arraystretch}{1.2}
\resizebox{\linewidth}{!}{
\begin{tabular}{l|cc|cc|ccccccccccccccc}
\toprule
& \multicolumn{2}{c|}{Train Set} & \multicolumn{2}{c|}{All Classes [\%]} & \multicolumn{15}{c}{Class-wise Intersection over Union (IoU) [\%]} \\
\cmidrule(lr){2-3} \cmidrule(lr){4-5} \cmidrule(lr){6-20}
Method & \multicolumn{1}{c|}{Manual GT} & SegFly-RGB & Acc. & FWmIoU & Road & Walkway & Dirt & Gravel & Grass & Vegetation & Tree & Grnd Obst. & Vehicle & Water & Building & Roof & Park. Lot & Constr. & Truck \\ 

\midrule
\rowcolor[HTML]{E6E6E6} 
\multicolumn{20}{l}{\textbf{Evaluation on Manual GT Test Set}} \\
\midrule
\multirow{2}{*}{UperNet~\cite{xiao2018upernet}} & \cmk & & $52.08$ & $52.46$ & $59.49$ & $\mathbf{40.05}$ & $\mathbf{17.47}$ & $\mathbf{36.73}$ & $\mathbf{59.49}$ & $06.44$ & $\mathbf{56.91}$ & $37.66$ & $\mathbf{63.12}$ & - & $71.82$ & $61.41$ & $34.98$ & $04.16$ & $\mathbf{01.77}$ \\
                         & & \cmk & $\mathbf{52.13}$ & $\mathbf{54.36}$ & $\mathbf{63.08}$ & $39.48$ & $09.60$ & $35.57$ & $57.15$ & $\mathbf{06.62}$ & $51.40$ & $\mathbf{49.18}$ & $61.86$ & - & $\mathbf{72.68}$ & $\mathbf{64.59}$ & $\mathbf{41.79}$ & $\mathbf{07.29}$ & $01.61$ \\
\midrule
\multirow{2}{*}{SegFormer~\cite{xie2021segformer}} & \cmk & & $50.05$ & $48.50$ & $55.15$ & $35.12$ & $05.10$ & $\mathbf{32.01}$ & $\mathbf{60.01}$ & $\mathbf{15.60}$ & $\mathbf{58.73}$ & $28.71$ & $\mathbf{66.67}$ & - & $65.67$ & $59.19$ & $\mathbf{24.05}$ & $08.01$ & $00.30$ \\
                           & & \cmk & $\mathbf{51.48}$ & $\mathbf{51.36}$ & $\mathbf{56.62}$ & $\mathbf{37.67}$ & $\mathbf{07.56}$ & $30.92$ & $59.66$ & $08.00$ & $56.08$ & $\mathbf{47.10}$ & $62.49$ & - & $\mathbf{69.25}$ & $\mathbf{65.12}$ & $23.14$ & $\mathbf{14.34}$ & $\mathbf{00.88}$ \\
\midrule
\multirow{2}{*}{Firefly (ours)} & \cmk & & $47.81$ & $45.81$ & $52.90$ & $39.71$ & $04.09$ & $26.58$ & $48.63$ & $07.78$ & $\mathbf{48.28}$ & $27.61$ & $46.11$ & - & $72.10$ & $55.13$ & $26.27$ & $05.14$ & $00.00$ \\
                               & & \cmk & $\mathbf{53.40}$ & $\mathbf{54.90}$ & $\mathbf{55.68}$ & $\mathbf{54.52}$ & $\mathbf{07.66}$ & $\mathbf{34.41}$ & $\mathbf{55.38}$ & $\mathbf{16.42}$ & $47.13$ & $\mathbf{52.91}$ & $\mathbf{62.09}$ & - & $\mathbf{77.62}$ & $\mathbf{66.40}$ & $\mathbf{38.62}$ & $\mathbf{06.10}$ & $\mathbf{00.06}$ \\
\midrule

\rowcolor[HTML]{E6E6E6} 
\multicolumn{20}{l}{\textbf{Evaluation on SegFly-RGB Test Set}} \\
\midrule
\multirow{2}{*}{UperNet~\cite{xiao2018upernet}} & \cmk & & $47.72$ & $43.72$ & $49.05$ & $\mathbf{39.50}$ & $\mathbf{18.33}$ & $24.89$ & $\mathbf{48.45}$ & $02.95$ & $\mathbf{47.95}$ & $35.73$ & $49.23$ & - & $59.23$ & $50.49$ & $\mathbf{29.56}$ & $\mathbf{02.07}$ & $00.26$ \\
                         & & \cmk & $\mathbf{48.38}$ & $\mathbf{46.73}$ & $\mathbf{49.74}$ & $35.99$ & $10.27$ & $\mathbf{26.85}$ & $47.41$ & $\mathbf{03.71}$ & $46.00$ & $\mathbf{56.17}$ & $\mathbf{55.23}$ & - & $\mathbf{64.22}$ & $\mathbf{55.45}$ & $26.23$ & $03.06$ & $\mathbf{01.79}$ \\
\midrule
\multirow{2}{*}{SegFormer~\cite{xie2021segformer}} & \cmk & & $46.02$ & $41.74$ & $\mathbf{47.12}$ & $34.29$ & $07.65$ & $20.45$ & $48.66$ & $\mathbf{06.86}$ & $46.31$ & $28.93$ & $53.03$ & - & $57.74$ & $48.38$ & $\mathbf{28.31}$ & $04.79$ & $00.45$ \\
                           & & \cmk & $\mathbf{48.01}$ & $\mathbf{44.46}$ & $44.40$ & $\mathbf{35.98}$ & $\mathbf{09.48}$ & $\mathbf{24.12}$ & $\mathbf{49.40}$ & $04.02$ & $\mathbf{47.01}$ & $\mathbf{52.92}$ & $\mathbf{57.01}$ & - & $\mathbf{61.53}$ & $\mathbf{54.50}$ & $17.81$ & $\mathbf{05.53}$ & $\mathbf{01.70}$  \\
\midrule
\multirow{2}{*}{Firefly (ours)} & \cmk & & $45.72$ & $42.34$ & $42.90$ & $42.66$ & $06.63$ & $13.46$ & $42.92$ & $03.19$ & $41.95$ & $28.25$ & $44.48$ & - & $71.45$ & $47.88$ & $26.73$ & $\mathbf{03.23}$ & $00.08$ \\
                               & & \cmk & $\mathbf{49.70}$ & $\mathbf{50.44}$ & $\mathbf{44.79}$ & $\mathbf{51.85}$ & $\mathbf{07.43}$ & $\mathbf{24.19}$ & $\mathbf{47.14}$ & $\mathbf{07.87}$ & $\mathbf{49.21}$ & $\mathbf{59.84}$ & $\mathbf{56.48}$ & - & $\mathbf{74.89}$ & $\mathbf{56.47}$ & $\mathbf{34.19}$ & $02.95$ & $\mathbf{01.02}$ \\
\bottomrule
\end{tabular}
}
\end{table}

\vspace{-1cm}
\begin{table}[h!]
\centering
\caption{RGB-T semantic segmentation results on SegFly-RGB-T test set.}
\label{tab_rgbt_results_supp}
\small
\renewcommand{\arraystretch}{1.2}
\resizebox{\linewidth}{!}{
\begin{tabular}{l | cc | cccccccccccccccc} 
\toprule
& \multicolumn{2}{c|}{All Classes [\%]} & \multicolumn{15}{c}{Class-wise Intersection over Union (IoU) [\%]} \\
\cmidrule(lr){2-3} \cmidrule(lr){4-18}  
Method & Acc. & FWmIoU & Road & Walkway & Dirt & Gravel & Grass & Vegetation & Tree & Grnd Obst. & Vehicle & Water & Building & Roof & Park. Lot & Constr. & Truck \\ 
\midrule
\rowcolor[HTML]{E6E6E6} 
\multicolumn{18}{l}{\textbf{SegFly-RGB-T Finetuning from Scratch}} \\
\hline
UperNet~\cite{xiao2018upernet}                       & $31.36$ & $27.82$ & $25.88$ & $\underline{31.22}$ & $00.38$ & $03.24$ & $27.04$ & $00.15$ & $\underline{14.33}$ & $32.72$ & $50.57$ & - & $26.09$ & $36.05$ & $\mathbf{22.43}$ & \underline{$00.32$} & $\underline{00.00}$ \\
SegFormer~\cite{xie2021segformer}                     & $\underline{33.26}$ & $\underline{28.33}$ & $\underline{32.49}$ & $28.63$ & $\underline{00.75}$ & $\mathbf{07.02}$ & $\mathbf{29.80}$ & $\mathbf{00.30}$ & $\mathbf{14.41}$ & $\underline{35.65}$ & $\underline{56.68}$ & - & $30.66$ & $34.30$ & \underline{$11.87$} & $\mathbf{00.63}$ & $\mathbf{00.01}$ \\
Firefly (ours)                & $\mathbf{37.48}$ & $\mathbf{35.20}$ & $\mathbf{47.61}$ & $\mathbf{36.82}$ & $\mathbf{01.45}$ & $\underline{05.09}$ & $\underline{29.71}$ & $00.03$ & $14.14$ & $\mathbf{45.37}$ & $\mathbf{57.20}$ & - & $\mathbf{45.30}$ & $\mathbf{40.22}$ & $08.39$ & $00.02$ & $\underline{00.00}$\\
\midrule
\rowcolor[HTML]{E6E6E6} 
\multicolumn{18}{l}{\textbf{SegFly-RGB Pretraining + SegFly-RGB-T Finetuning}} \\
\hline
UperNet~\cite{xiao2018upernet}                                 & $33.48$ & $32.21$ & $32.81$ & \underline{$33.10$} & \underline{$00.67$} & $\mathbf{04.28}$ & \underline{$30.91$} & $00.08$ & $13.28$ & \underline{$37.80$} & $47.66$ & - & \underline{$33.80$} & \underline{$42.70$} & \underline{$18.71$} & $00.21$ & $\mathbf{00.07}$ \\
SegFormer~\cite{xie2021segformer}                               & \underline{$34.26$} & $30.66$ & \underline{$37.15$} & $32.10$ & $00.50$ & \underline{$03.47$} & $29.55$ & $\mathbf{00.29}$ & $\mathbf{16.16}$ & $36.96$ & \underline{$57.30$} & - & $29.40$ & $37.53$ & $\mathbf{22.35}$ & $\mathbf{00.80}$ & $00.00$\\
Firefly-RGB (ours)                      & $\mathbf{37.67}$ & $\mathbf{37.38}$ & $\mathbf{47.42}$ & $\mathbf{36.49}$ & $\mathbf{01.52}$ & $02.83$ & $\mathbf{31.01}$ & \underline{$00.27$} & \underline{$13.99$} & $\mathbf{48.66}$ & $\mathbf{59.31}$ & - & $\mathbf{49.83}$ & $\mathbf{42.93}$ & $10.84$ & \underline{$00.55$} & $00.00$ \\
\midrule
\rowcolor[HTML]{E6E6E6} 
\multicolumn{18}{l}{\textbf{SegFly-RGB Pretraining + SegFly-RGB-T Adaptation + SegFly-RGB-T Finetuning}} \\
\hline
UperNet~\cite{xiao2018upernet} + Rein~\cite{wei2024rein}                       & $34.74$ & $31.98$ & $39.56$ & \underline{$30.81$} & $00.92$ & $03.44$ & $\mathbf{32.62}$ & $00.09$ & $12.77$ & $39.23$ & $55.41$ & - & $31.12$ & $39.86$ & $\mathbf{19.91}$ & \underline{$00.46$} & $00.00$ \\
SegFormer~\cite{xie2021segformer} + Rein~\cite{wei2024rein}                      & \underline{$35.75$} & \underline{$33.48$} & \underline{$42.31$} & $29.19$ & \underline{$00.39$} & $\mathbf{05.19}$ & $30.88$ & \underline{$00.10$} & \underline{$14.91$} & \underline{$44.71$} & \underline{$56.17$} & - & \underline{$33.34$} & \underline{$42.49$} & \underline{$17.62$} & $00.28$ & $00.00$\\
Firefly-RGB-T (ours)                 & $\mathbf{41.06}$ & $\mathbf{43.83}$ & $\mathbf{46.69}$ & $\mathbf{51.08}$ & $\mathbf{07.11}$ & \underline{$05.16$} & \underline{$32.42$} & $\mathbf{00.85}$ & $\mathbf{17.20}$ & $\mathbf{54.90}$ & $\mathbf{56.92}$ & - & $\mathbf{62.69}$ & $\mathbf{49.72}$ & $14.72$ & $\mathbf{00.51}$ & $00.00$\\
\bottomrule
\end{tabular}
}
\end{table}

\begin{table}[h!]
\centering
\caption{Vision Foundation Models on SegFly-RGB and SegFly-RGB-T.}
\label{tab_vision_foundation_models_supp}
\small
\renewcommand{\arraystretch}{1.3}
\resizebox{\linewidth}{!}{
\begin{tabular}{c|c|c|cc|ccccccccccccccc}
\toprule
& Eval & Class & \multicolumn{2}{c|}{All Classes [\%]} & \multicolumn{15}{c}{Class-wise Intersection over Union (IoU) [\%]} \\
\cline{4-5} \cline{6-20}
Method & Type & Capacity & Acc. & FWmIoU & Road & Walkway & Dirt & Gravel & Grass & Vegetation & Tree & Grnd Obst. & Vehicle & Water & Building & Roof & Park. Lot & Constr. & Truck \\ 

\midrule
\rowcolor[HTML]{E6E6E6} 
\multicolumn{20}{l}{\textbf{SegFly-RGB Semantic Segmentation}} \\
\hline
\multirow{1}{*}{CatSeg~\cite{cho2024catseg}} & Zero-Shot & Open Vocab. & $47.59$ & $28.63$ & $37.38$ & $28.01$ & $14.56$ & $0.29$ & $54.35$ & $2.25$ & $\mathbf{61.96}$ & $0.00$ & $53.89$ & - & $49.40$ & $26.28$ & $0.82$ & $0.65$ & $\mathbf{20.02}$ \\
(CVPR 2024) & Fine-Tuned & Open Vocab. & $\mathbf{70.64}$ & $\mathbf{55.44}$ & $\mathbf{45.57}$ & $\mathbf{52.30}$ & $\mathbf{17.50}$ & $\mathbf{22.45}$ & $\mathbf{56.61}$ & $\mathbf{10.67}$ & 54.48 & $\mathbf{65.12}$ & $\mathbf{58.65}$ & - & $\mathbf{79.23}$ & $\mathbf{63.02}$ & $\mathbf{45.67}$ & $\mathbf{5.27}$ & $12.10$  \\

\midrule

\rowcolor[HTML]{E6E6E6} 
\multicolumn{20}{l}{\textbf{SegFly-RGB-T Semantic Segmentation}} \\
\hline
\multirow{1}{*}{AnyThermal~\cite{mahes2026anythermal}} & Zero-Shot & \multicolumn{1}{l|}{10 (CART~\cite{lee2024cart})} & $24.94$ & $7.73$ & $21.82$ & $0.00$ & $0.00$ & $0.00$ & $0.00$ & $0.00$ & $\mathbf{10.04}$ & $0.00$ & $32.91$ & - & $25.96$ & $0.00$ & $0.00$ & $0.00$ & $0.00$ \\
(ICRA 2026) & Fine-Tuned & 15 (SegFly, ours) & $\mathbf{40.50}$ & $\mathbf{28.26}$ & $\mathbf{38.11}$ & $\mathbf{25.95}$ & $\mathbf{7.72}$ & $\mathbf{3.49}$ & $\mathbf{26.83}$ & $\mathbf{0.14}$ & $8.02$ & $\mathbf{40.66}$ & $\mathbf{53.48}$ & - & $\mathbf{31.32}$ & $\mathbf{30.69}$ & $\mathbf{16.63}$ & $\mathbf{0.05}$ & $\mathbf{0.06}$ \\
\bottomrule
\end{tabular}
}
\end{table}

\section{Benchmark Implementation Details}

\noindent \textbf{Evaluation Metrics.}\;
For all benchmarks, we report three standard semantic segmentation metrics.
(i) Global pixel accuracy (Acc.) measures the proportion of correctly classified pixels over all labeled pixels.
(ii) Class-wise intersection-over-union (IoU) is computed separately for each of the 15 semantic classes.
(iii) Frequency-weighted mean IoU (FWmIoU) aggregates performance by averaging per-class IoUs weighted by the empirical pixel-frequency of each class. Unlabeled pixels are treated as an ignore index and are excluded from all metric computations, while FWmIoU is always computed over the 15 foreground classes only.\\

\noindent \textbf{Model-Agnostic Details.}\;
All models are implemented in PyTorch and trained with mixed-precision (fp16) on a single NVIDIA H100 GPU with 96 GB memory. We use the AdamW~\cite{adam} optimizer with standard $\beta_1 = 0.9$, $\beta_2 = 0.99$, and a batch size of 16. For the thermal modality, inputs are normalized using a dataset-level mean and standard deviation computed over all training images. For every method, we select the final checkpoint as the epoch with the highest validation FWmIoU.\\

\noindent \textbf{SegFormer Details.}\;
For SegFormer~\cite{xie2021segformer}, we use the official implementation with the MiT-B3 backbone. The model has \SI{47.23}{M} trainable parameters and is trained with a standard pixel-wise cross-entropy loss. Following the training setup proposed in the original work and considering dataset sizes, we train for 128 epochs with an initial learning rate of $6\times 10^{-5}$. Training SegFormer on SegFly takes approximately 64 hours.\\

\noindent \textbf{UPerNet Details.}\;
For UPerNet~\cite{xiao2018upernet}, we adopt the official configuration with a Swin-Small backbone, resulting in \SI{81.16}{M} trainable parameters. As for SegFormer, we use a standard cross-entropy loss. In line with the proposed training regime, we train for 128 epochs with an initial learning rate of $6\times 10^{-5}$, matching the effective training length of the original setup. On SegFly, UPerNet training requires about 81 hours.\\

\noindent \textbf{Firefly Details.}\;
Firefly builds on a frozen DINOv3 ViT-Base/16 encoder (\SI{85.60}{M} parameters) and a lightweight segmentation head. The head is a three-layer point-wise MLP with ReLU activations in all hidden layers and a final linear layer that predicts per-class logits, adding \SI{2.36}{M} parameters. For RGB experiments, we train only the MLP head with a standard Dice loss~\cite{Milletari_2016_diceloss_vnet} for 40 epochs  using an initial learning rate of $1\times 10^{-4}$, which results in a total training time of 23 hours. For thermal semantic segmentation, we extend the encoder with Rein adapters (\SI{2.50}{M} parameters) following the original design and use them only for the thermal branch. During RGB–T domain adaptation and thermal fine-tuning, the backbone remains frozen while the Rein adapters and the thermal MLP head are optimized with the same Dice loss and learning rate as above.\\

\noindent \textbf{Compute Cost.}\;
UPerNet/SegFormer/Firefly need 402/127/141 GFLOPs in the RGB domain, and 413/134/156 GFLOPs in the thermal domain, respectively.

\end{document}